\documentclass[twocolumn]{autart} 
\usepackage{graphicx}          
\usepackage{mathrsfs}
\usepackage{amssymb,setspace}
\usepackage{amsmath,bm}
\usepackage{amsfonts}
\usepackage{epsfig}
\usepackage{enumerate}
\usepackage{graphicx}
\usepackage{subfigure}
\usepackage{arydshln}
\usepackage{lipsum}
\usepackage{float}
\usepackage{galois}
\usepackage{color, soul}
\usepackage{float}
\usepackage{algorithm}  
\usepackage{algorithmicx} 
\usepackage[noend]{algpseudocode}
\usepackage{romannum}
\usepackage{url}
\usepackage{cite}
\usepackage{textcomp}
\usepackage{etex}
\usepackage{amsmath,amssymb,pstricks,color,dsfont, bm, amsfonts}
\let\theoremstyle\relax
\usepackage{amsthm}
\usepackage[round,authoryear,sort,comma]{natbib}
\usepackage{mathtools, cuted}
\usepackage{algorithm, algpseudocode, xspace}
\usepackage{epstopdf}
\usepackage{multirow, multicol}
\usepackage{enumitem}

\usepackage{url}
\usepackage{hyperref}
\usepackage{picins}

\newenvironment{autoBiography}[3]{\parpic{\includegraphics[width=1in,clip,keepaspectratio]{#1}}\noindent {\bf{#2}} #3}

\allowdisplaybreaks

\newtheorem{Def}{Definition}
\newtheorem{Lem}{Lemma}
\newtheorem{Asu}{Assumption}
\newtheorem{Thm}{Theorem}
\newtheorem{Rem}{Remark}

\begin{document}

\begin{frontmatter}

\title{Byzantine-resilient federated online learning for Gaussian process regression} 

\thanks[footnoteinfo]{This paper was not presented at any IFAC 
meeting. Corresponding author Minghui Zhu. Tel. +1-814-865-5315. }

\author[PSU]{Xu Zhang}\ead{xxz313@psu.edu},    
\author[VT]{Zhenyuan Yuan }\ead{zyuan18@vt.edu},               
\author[PSU]{Minghui Zhu$^\star$}\ead{muz16@psu.edu}  

\address[PSU]{School of Electrical Engineering and Computer Science, The Pennsylvania State University, University Park, PA 16802 USA}
\address[VT]{Virginia Tech Transportation Institute, Virginia Polytechnic Institute and State University, Blacksburg, VA 24061 USA}


\begin{abstract}                          
In this paper, we study Byzantine-resilient federated online learning for Gaussian process regression (GPR). We develop a Byzantine-resilient federated GPR algorithm that allows a cloud and a group of agents to collaboratively learn a latent function {{and improve the learning performances}} where some agents exhibit Byzantine failures, i.e., arbitrary and potentially adversarial behavior. Each agent-based local GPR sends potentially compromised local predictions to the cloud, and the cloud-based aggregated GPR computes a global model by a Byzantine-resilient product of experts aggregation rule. Then the cloud broadcasts the current global model to all the agents. Agent-based fused GPR refines local predictions by fusing the received global model with that of the agent-based local GPR. Moreover, we quantify the learning accuracy improvements of the agent-based fused GPR over the agent-based local GPR. Experiments on a toy example and two medium-scale real-world datasets are conducted to demonstrate the performances of the proposed algorithm. 

\end{abstract}

\end{frontmatter}

\section{Introduction}
Cloud computing has garnered considerable attention for decades, and its advantages include unlimited data storage, powerful data processing, scalability and cost efficiency. Since machine learning is a data-intensive and time-intensive endeavor, cloud-based machine learning emerges and performs in a fast, scalable and affordable way  \citep{HGGG2012, KPA2015, MG2009}. However, as the amount of collected data significantly increases, cloud-based machine learning framework may incur large communication overheads. For example, a modern vehicle can generate hundreds of gigabytes of data in several hours, which is a huge burden for data transmission and storage \citep{MRC2016}. Moreover, this framework faces privacy risk during data transmission to the cloud, as adversaries can infer private information of data owners from collected raw data. For example, an adversary notices that \emph{Tom} visits a hospital every day, and can discover that \emph{Tom} or one of his relatives or friends has a health problem \citep{J2012}. To address the above challenges, federated learning provides a promising paradigm  and two superiorities: 1) only a finite number of model parameters are shared and their size is much smaller than that of raw data; 2) data samples are kept by data owners and local machine learning models are trained without sharing raw data with each other \citep{yuan2024federated, KJM2016, MBM2017, YQLY2019}. Hence federated learning reduces communication overheads on the network and preserves privacy of owners' raw data. Despite its promise, federated learning faces significant security challenges. For example, Byzantine attacks compromise data owners such that they deviate from expected training processes.

\emph{Literature review.} Attacks on federated learning can be categorized into two classes: data poisoning attacks and model poisoning attacks. 

Data poisoning attacks aim at injecting corrupted samples into the training dataset to train local models in hopes of producing incorrect model parameters which are sent to the cloud \citep{VR2021}. Paper \cite{VSM2020} uses two image classification datasets to experimentally demonstrate that data poisoning attacks can significantly decrease classification accuracy. This paper further proposes a defense, which can identify malicious agents. Given the fact that malicious agents' updates deviate from benign agents' updates, a defense system is proposed in \cite{YTG2020} to  detect data poisoning attacks by computing similarity scores.  

Data poisoning attacks compromise the integrity of training dataset. On the other hand, model poisoning attacks aim to compromise the integrity of learning process at the training stage and directly manipulate local model parameters. They can be referred to as Byzantine attacks. Papers \cite{BEG2017}, \cite{YLJ2017}, \cite{DS2020}, {{\cite{SRNRJ2022}}}, {{\cite{ESP2022}}}, {{\cite{SPLM2021}}}, {{\cite{SPLM2022}}}, \cite{JBA2021} and \cite{YCRB2018} study Byzantine-resilient federated learning where the cloud and a group of agents aim to collaboratively train a global model using agents' data despite Byzantine attacks. Paper \cite{BEG2017} proposes \emph{Krum} as the aggregation rule where the cloud is able to select a gradient such that the overall distance between this gradient to a fixed number of nearby gradients is minimal. Paper \cite{JBA2021} develops an algorithm which is based on an integrated stochastic quantization, geometric median based outlier detection, and secure model aggregation. Paper \cite{YCRB2018} analyzes two robust distributed gradient descent algorithms based on median and trimmed mean operations, respectively.  Paper \cite{DS2020} proposes a high-dimensional robust mean estimation algorithm at the cloud to combat the adversary. In paper \cite{YLJ2017}, the cloud partitions all the received local gradients into batches and computes the mean of each batch. After that, the cloud computes the geometric median of the batch means. The above set of papers can guarantee that the estimation error of the parameter aggregated by the cloud is upper bounded and determined by the number of Byzantine agents. Moreover, \cite{BEG2017} and  \cite{JBA2021} show that the sequence of the gradients converges almost surely to zero, and \cite{YLJ2017}, \cite{DS2020}, \cite{YCRB2018} show that the parameter converges to a neighborhood of the optimum at an exponential rate for strongly-convex objective functions.
However, the aforementioned all papers only consider deep neural networks (DNN) as the learning model and are limited to static data and off-line learning.

Gaussian process regression (GPR) is an efficient non-parametric method for on-line learning \citep{KDAN2013, HYXJ2020, MXB2016, CC2006, XHEC2013}. Distributed implementation of GPR over peer-to-peer networks has been studied in recent papers. Paper \cite{JKYP2015} develops Gaussian process decentralized data fusion and active sensing algorithms for cooperative prediction of traffic phenomenon with a fleet of vehicles. Paper \cite{DJCDH2020} presents a distributed GPR algorithm using Karhunen-Lo\'{e}ve expansion and an average consensus protocol, which allows multiple robots to collaboratively explore an unknown area. Paper \cite{THJ2016} proposes a decentralized computation and centralized data fusion algorithm using Gaussian processes, and this algorithm enables robots to visit the most important location for active sensing. In paper \cite{ZM2021}, distributed GPR algorithm enables a group of agents to jointly learn a common static latent function on-line via limited inter-agent communication. Notice that, none of the aforementioned papers considers {{federated GPR}}, and security issues induced by Byzantine attacks have not been thoroughly studied yet.


In our conference paper \cite{XZM2022}, we consider Byzantine-resilient federated learning for streaming data using GPR where a cloud and a group of agents aim to collaboratively learn a latent function despite some agents exhibit Byzantine failures.  The prediction error is positively related to the number of Byzantine agents and that of the agents with extreme values. {{We quantify the lower and upper bounds of the predictive variance}}. In the experiments, a toy example and two
medium-scale real-world datasets are used to
demonstrate that the Byzantine-resilient GPR algorithm
is robust to Byzantine attacks.

\emph{Contributions and distinctions.} This paper extends the conference paper {{\cite{XZM2022}}} through developing a new prediction fusion algorithm for the agent-based fused GPR. The fusion algorithm chooses the better predictions between those of the agent-based local GPR and the cloud-based aggregated GPR for each test input. In theoretical analysis, we first derive the upper bounds on the prediction errors of the agent-based fused GPR for the benign agents, and then quantify the improvements of the learning performance of the agent-based fused GPR over the agent-based local GPR by directly comparing their mean squared errors. We also derive the bounds on the predictive variance of the agent-based fused GPR. Compared with \cite{XZM2022}, we derive a tighter upper bound of the prediction error of the cloud-based aggregated GPR. This paper conducts {{additional experiments}} to demonstrate the improvements of the agent-based fused GPR over the agent-based local GPR.

%

\emph{Notations and notions.}
Throughout the paper, we use lower-case letters, e.g, $a$, to denote scalars, bold letters, e.g., $\boldsymbol{a}$, to denote vectors; upper-case letters, e.g., $A$, to denote matrices, calligraphic letters, e.g., $\mathcal{A}$, to denote sets, and bold calligraphic letters, e.g., $\boldsymbol{\mathcal{A}}$, to denote spaces. Denote $I_n\in\mathbb{R}^{n\times n}$ the $n$-by-$n$ identity matrix. We denote by $\mathbb{R}$ the set of real numbers. The set of non-negative real numbers is denoted by $\mathbb{R}_+$, and the set of positive real numbers is denoted by $\mathbb{R}_{++}$. The cardinality of set $\mathcal{A}$ is denoted by $|\mathcal{A}|$. Define the distance metric $D(\boldsymbol{z},\boldsymbol{z}')\triangleq \|\boldsymbol{z}-\boldsymbol{z}'\|$, and the point to set distance as $D(\boldsymbol{z},\mathcal{Z})\triangleq\inf_{\boldsymbol{z}'\in\mathcal{Z}}D(\boldsymbol{z},\boldsymbol{z}')$. Define  $\text{proj}(\boldsymbol{z},\mathcal{Z})\triangleq \{\boldsymbol{z}'\in\boldsymbol{\mathcal{Z}}|D(\boldsymbol{z},\boldsymbol{z}')=D(\boldsymbol{z},\mathcal{Z})\}$ the projection set of point $\boldsymbol{z}$ onto set $\mathcal{Z}$. Denote the supremum of a function $\eta:\boldsymbol{\mathcal{Z}}\rightarrow\mathbb{R}$ as $\|\eta\|_{\infty}\triangleq\sup_{\boldsymbol{z}\in\boldsymbol{\mathcal{Z}}}|\eta(\boldsymbol{z})|$. We denote by $\mathcal{O}(g(t))$ the limiting behavior of some function $f(t)$ if $\lim_{t\rightarrow\infty}\frac{f(t)}{g(t)} = a$ for some constant $a>0$. We use $(\cdot)^{[i]}$ to distinguish the local values of agent $i$, and $(\cdot)^{\max}$ $((\cdot)^{\min})$ denote the maximum (minimum) of the local values, e.g., $a^{\max}\triangleq\max_{i\in\mathcal{V}}a^{[i]}$.

We give formal definitions of sub-Gaussian random variables with parameter $\sigma$ and Gaussian process.
	\begin{Def} \citep{Martin2019} \label{Def:1}
	A random variable $X$ with mean $\mu\triangleq\mathbb{E}\left[X\right]$ is sub-Gaussian (denoted by $X\sim subG(\sigma^2)$) if there exists a positive number $\sigma$ such that 
	\begin{align}\label{eq:1}
	\mathbb{E}\left[\exp(\lambda(X-\mu))\right]\le \exp\left(\frac{{\sigma}^2\lambda^2}{2}\right),\quad \forall\lambda\in\mathbb{R}.
	\end{align}
\end{Def}
\begin{Def}  \citep{CC2006}
	A Gaussian process is a collection of random variables, any finite number of which have a joint Gaussian distribution.
\end{Def}
\section{Preliminary}
This section provides basic knowledge of GPR. The presentation strictly follows \cite{CC2006}. Let $\eta:\boldsymbol{\mathcal{Z}}\rightarrow\boldsymbol{\mathcal{Y}}$ be the target function, where $\boldsymbol{\mathcal{Z}}\subseteq\mathbb{R}^{n_z}$ and $\boldsymbol{\mathcal{Y}}\subseteq\mathbb{R}$. Given input $\boldsymbol{z}(t)\in\boldsymbol{\mathcal{Z}}$ at time $t$, the corresponding output is given by 
\begin{align}\label{eq:2}
y(t) = \eta(\boldsymbol{z}(t))+e(t), \quad e(t)\sim\mathcal{N}(0,\sigma_e^2),
\end{align}
where $e(t)$ is the Gaussian measurement noise. Let training data be in the form $\mathcal{D}\triangleq\{(\boldsymbol{z}(t),y(t))|t=1,\ldots,n_s\}$. GPR aims to infer the values of the latent function $\eta$ over a set of test inputs $\mathcal{Z}_*\subset \boldsymbol{\mathcal{Z}}$ using $\mathcal{D}$.

Define kernel function ${{{\rm ker}}}:\mathbb{R}^{n_z}\times \mathbb{R}^{n_z} \rightarrow \mathbb{R}$ that is symmetric and positive semi-definite. 
Assume that the latent function $\eta$ is sampled from a GP prior specified by mean zero and kernel function {{$\rm ker$}}. Then the training outputs $\boldsymbol{y}$ and the test outputs denoted by $\boldsymbol{\eta}_*$ are jointly Gaussian distributed as:
\begin{align*}
\left[\begin{matrix}
\boldsymbol{y} \\
\boldsymbol{\eta}_*
\end{matrix}\right] \sim \mathcal{N} (\boldsymbol{0}, \left[\begin{matrix}
{{\rm ker}}(\mathcal{Z},\mathcal{Z})+\sigma_e^2I_{n_s} & {{\rm ker}}(\mathcal{Z},\mathcal{Z}_*) \\
{{\rm ker}}(\mathcal{Z}_*,\mathcal{Z}) & {{\rm ker}}(\mathcal{Z}_*,\mathcal{Z}_*)
\end{matrix}\right]),
\end{align*}
where ${{\rm ker}}(\mathcal{Z},\mathcal{Z}_*)$ returns a matrix such that the entry is evaluated at the pair of training and test points. The notations ${{\rm ker}}(\mathcal{Z},\mathcal{Z})$, ${{\rm ker}}(\mathcal{Z}_*,\mathcal{Z})$ and ${{\rm ker}}(\mathcal{Z}_*,\mathcal{Z}_*)$ are defined in an analogous way.

Utilizing identities of joint Gaussian distribution (see page 200 in \cite{CC2006}), GPR makes predictions of $\boldsymbol{\eta}$ on $\mathcal{Z}_*$ based on dataset $\mathcal{D}$ as $\boldsymbol{\eta}_*\sim\mathcal{N}(\boldsymbol{\mu}_{\mathcal{Z}_*|\mathcal{D}},\Sigma_{\mathcal{Z}_*|\mathcal{D}})$, where
\begin{align}\label{eq:3}
\boldsymbol{\mu}_{\mathcal{Z}_*|\mathcal{D}}&\triangleq {{\rm ker}}(\mathcal{Z}_*,\mathcal{Z})\mathring{k}(\mathcal{Z},\mathcal{Z})^{-1}\boldsymbol{y}, \nonumber \\
\Sigma_{\mathcal{Z}_*|\mathcal{D}}&\triangleq {{\rm ker}}(\mathcal{Z}_*,\mathcal{Z}_*)-{{\rm ker}}(\mathcal{Z}_*,\mathcal{Z})\mathring{k}(\mathcal{Z},\mathcal{Z})^{-1}{{\rm ker}}(\mathcal{Z},\mathcal{Z}_*),
\end{align}
with $\mathring{k}(\mathcal{Z},\mathcal{Z})\triangleq {{\rm ker}}(\mathcal{Z},\mathcal{Z})+\sigma_e^2I_{n_s}$. We refer (\ref{eq:3}) as full GPR. If $y$ is multi-dimensional, GPR is performed for each element.
%

\section{Problem formulation}
\subsection{Network model}
Consider a network consisting of $n$ agents and a cloud. We denote the agent set by $\mathcal{V}\triangleq\{1,\ldots,n\}$. At each time instant $t$, the observation model of agent $i$ is given as
\begin{align}\label{eq:4}
y^{[i]}(t) = \eta(\boldsymbol{z}^{[i]}(t))+e^{[i]}(t)
\end{align}
where $\boldsymbol{z}^{[i]}(t)\in\boldsymbol{\mathcal{Z}}\subseteq\mathbb{R}^{n_z}$ is the input of $\eta$, $y^{[i]}(t)\in\boldsymbol{\mathcal{Y}}\subseteq\mathbb{R}$ is the observation of the agent, and $e^{[i]}(t)$ is independent Gaussian noise with zero mean and variance $(\sigma_e^{[i]})^2$. The agents can communicate with the cloud but they cannot communicate with each other. Pickup count prediction of the mobility-on-demand system with a fleet of robotic vehicles can be such an application \citep{JKYP2015}.

\subsection{Attack model}
All communications between the cloud and the agents are attack-free. Some agents exhibit Byzantine failures and behave arbitrarily during learning process, e.g., disobey the protocols
and send arbitrary messages. For ease of presentation, we call these agent by Byzantine agents and others by benign agents. We assume that an $\alpha$ fraction of the agents are Byzantine agents, and the remaining $1-\alpha$ fraction are benign agents. Note that $\alpha n$ and $(1-\alpha) n$ are integers. We denote $\mathcal{B}\subset\mathcal{V}$ as the set of Byzantine agents. Both the cloud and the benign agents are unaware of the identities of Byzantine agents, but the cloud knows that $\frac{1}{4}$ is an upper bound of $\alpha$. 

\subsection{Objective}
The objective of this paper is to design an algorithm, which allows the benign agents to learn a common latent function $\eta$ {{and enhance the learning performances}} using streaming data $({\boldsymbol{z}^{[i]}(t), y^{[i]}(t)})$ despite less than one quarter of the agents are Byzantine agents.

\section{Byzantine-resilient federated GPR}
\begin{figure}[htbp]
	\centering
	\includegraphics[width=0.45\textwidth]{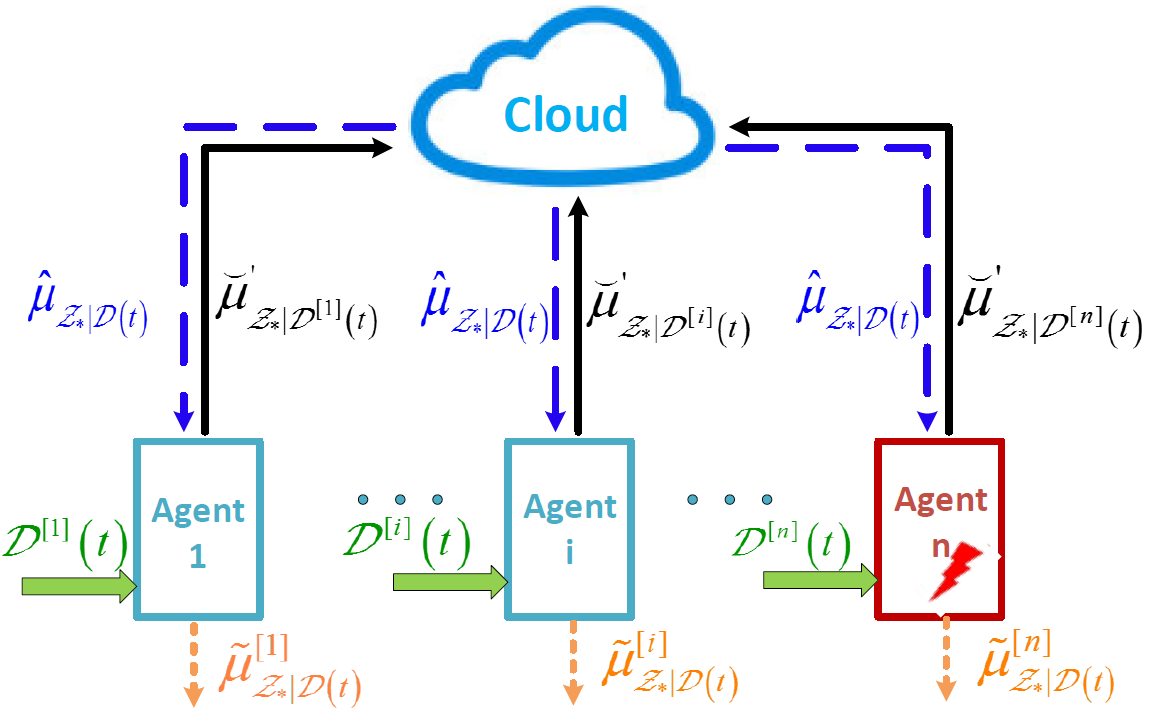}
	\caption{Diagram of Byzantine-resilient federated GPR.} \label{Fig:0}
\end{figure}

This section develops a Byzantine-resilient federated GPR algorithm in Fig. \ref{Fig:0} to solve the problem formulated in Section 3. This algorithm consists of three modules: agent-based local GPR, cloud-based aggregated GPR and agent-based fused GPR.  

\subsection{Algorithm statement}
First at each time instant $t$, each agent $i\in\mathcal{V}$ makes its own prediction of $\eta$ through the agent-based local GPR over a given set of test points $\mathcal{Z}_*\subseteq\boldsymbol{\mathcal{Z}}$ using local streaming dataset $\mathcal{D}^{[i]}(t)\triangleq(\mathcal{Z}^{[i]}(t),\boldsymbol{y}^{[i]}(t))$ with local input data $\mathcal{Z}^{[i]}(t) \triangleq \{\boldsymbol{z}^{[i]}(1),\ldots,\boldsymbol{z}^{[i]}(t)\}$ and output $\boldsymbol{y}^{[i]}(t)\triangleq[y^{[i]}(1),\ldots,y^{[i]}(t)]^{\rm T}$. Then the benign agents send correct local predictions to the cloud and the Byzantine agents instead send arbitrary values. Second, the cloud collects all the local predictions and constructs two agent sets by trimming a fraction of outliers with respect to the predictive means and variances. Then the cloud uses a Byzantine-resilient PoE aggregation rule to compute a global prediction, and broadcasts the global prediction to all the agents. Third, the agents refine the predictions on $\mathcal{Z}_*$ by fusing the predictions from the cloud-based aggregated GPR with those from the agent-based local GPR. Each agent only communicates local predictions with the cloud, and does not share its local streaming data with the cloud and other agents. The formal description of Byzantine-resilient federated GPR is given in Algorithm \ref{alg:1}. Next we elaborate on the three modules.

\begin{algorithm}  
	\caption{Byzantine-resilient federated GPR}  
	\label{alg:1}  
	\begin{algorithmic}[1] 
		\State \textbf{Initialization}: $\mathcal{D}(0)=\emptyset$. 
		\For{$t=1,2,\ldots$}

		\{Agent-based local GPR\}
		\For{$i\in\mathcal{V}$}
		
		\State $\mathcal{D}^{[i]}(t)=\mathcal{D}^{[i]}(t-1)\bigcup(\boldsymbol{z}^{[i]}(t),y^{[i]}(t))$ 
		
		\State $\check{\boldsymbol{\mu}}'_{\mathcal{Z}_*|\mathcal{D}^{[i]}(t)},\check{\boldsymbol{\sigma}}'^2_{\mathcal{Z}_*|\mathcal{D}^{[i]}(t)}=\text{lGPR}(\mathcal{D}^{[i]}(t))\quad\qquad$
		\EndFor

		\{{Cloud-based aggregated GPR}\}
		\State $\hat{\boldsymbol{\mu}}_{\mathcal{Z}_{*}|\mathcal{D}(t)}, \hat{\boldsymbol{\sigma}}^2_{\mathcal{Z}_{*}|\mathcal{D}(t)}
		\text{= cGPR}(\check{\boldsymbol{\mu}}'_{\mathcal{Z}_{*}|\mathcal{D}^{[i]}(t)},\check{\boldsymbol{\sigma}}'^2_{\mathcal{Z}_{*}|\mathcal{D}^{[i]}(t)})$
		
		\{{Agent-based fused GPR}\}
		\For{$i\in\mathcal{V}$}
		
		\State
		$\tilde{\boldsymbol{\mu}}^{[i]}_{\mathcal{Z}_*|\mathcal{D}(t)}, (\tilde{\boldsymbol{\sigma}}^{[i]}_{\mathcal{Z}_*|\mathcal{D}(t)})^2$$
		=\text{fGPR}(\check{\boldsymbol{\mu}}'_{\mathcal{Z}_*|\mathcal{D}^{[i]}(t)},\check{\boldsymbol{\sigma}}'^2_{\mathcal{Z}_*|\mathcal{D}^{[i]}(t)},\hat{\boldsymbol{\mu}}_{\mathcal{Z}_{*}|\mathcal{D}(t)},\hat{\boldsymbol{\sigma}}^2_{\mathcal{Z}_{*}|\mathcal{D}(t)})$
		
		\EndFor
		
		\EndFor
	\end{algorithmic}  
\end{algorithm} 

\begin{algorithm}  
	\caption{Agent-based local GPR  } 
	\label{Alg:2}  
	\begin{algorithmic}[1] 
		\State lGPR$(\mathcal{D}^{[i]}(t))$
		\State $\mathcal{Z}^{[i]}(t) = \{\boldsymbol{z}^{[i]}(1),\ldots,\boldsymbol{z}^{[i]}(t)\}$ 
		\State $\boldsymbol{y}^{[i]}(t)=[y^{[i]}(1),\ldots,y^{[i]}(t)]^{\rm T}$
		\For{$\boldsymbol{z}_*\in\mathcal{Z}_*$}
		
		\State Choose $z_*^{[i]}(t)~\in~\text{proj}(\boldsymbol{z}_*,\mathcal{Z}^{[i]}(t))$ 
		
		\State Compute
		\begin{align*}
		\check{{\mu}}'_{\boldsymbol{z}_{*}|\mathcal{D}^{[i]}(t)}\leftarrow\left\{\begin{array}{ll}
		\check{{\mu}}_{\boldsymbol{z}_{*}|\mathcal{D}^{[i]}(t)} & \text{Benign agent}, \\
		\star & \text{Byzantine agent}
		\end{array}\right.
		\end{align*} 
		\State Compute
		\begin{align*}
		\check{{\sigma}}'^2_{\boldsymbol{z}_{*}|\mathcal{D}^{[i]}(t)}\leftarrow\left\{\begin{array}{ll}
		\check{{\sigma}}^2_{\boldsymbol{z}_{*}|\mathcal{D}^{[i]}(t)} & \text{Benign agent}, \\
		\star & \text{Byzantine agent}
		\end{array}\right.
		\end{align*}
		\State Send $\check{{\mu}}'_{\boldsymbol{z}_{*}|\mathcal{D}^{[i]}(t)}$ and $\check{{\sigma}}'^2_{\boldsymbol{z}_{*}|\mathcal{D}^{[i]}(t)}$ to the cloud
		\EndFor
		
		\State \Return $\check{\boldsymbol{\mu}}'_{\mathcal{Z}_*|\mathcal{D}^{[i]}(t)}$, $\check{\boldsymbol{\sigma}}'^2_{\mathcal{Z}_*|\mathcal{D}^{[i]}(t)}$
	\end{algorithmic}  
\end{algorithm} 

\subsubsection{Agent-based local GPR}
To reduce computational complexity, the agent-based local GPR performs Nearest-neighbor GPR (NNGPR)  to predict for each test point $\boldsymbol{z}_*\in\mathcal{Z}_*$. Instead of feeding the whole training dataset to GPR, the agent-based local GPR only uses the nearest input denoted by $\boldsymbol{z}_*^{[i]}(t)~\in~\text{proj}(\boldsymbol{z}_*,\mathcal{Z}^{[i]}(t))$ and its corresponding output $y_{\boldsymbol{z}_*^{[i]}(t)}^{[i]}$ to compute the local predictions  \citep{ZM2021}. Notice that computation complexity of the hyperparameter tuning for NNGPR is $\mathcal{O}(t^3)$. In fact, the computation complexity of the hyperparameter tuning can be reduced to $\mathcal{O}(1)$ by using recursive hyperparameter tuning  \citep{HMF2004}. NNGPR in prediction has computation complexity equivalent to nearest-neighbor search, whose worst case complexity is $\mathcal{O}(t)$, while full GPR in (\ref{eq:3}) has computation complexity $\mathcal{O}(t^3)$  \citep{AGN2014}.

Denote $\check{{\mu}}'_{\boldsymbol{z}_{*}|\mathcal{D}^{[i]}(t)}$ and $\check{\sigma}'^2_{\boldsymbol{z}_{*}|\mathcal{D}^{[i]}(t)}$ the predictive mean and variance from the agent-based local GPR. The predictive mean and variance of the benign agents are written in the following forms $\check{{\mu}}'_{\boldsymbol{z}_{*}|\mathcal{D}^{[i]}(t)}=\check{{\mu}}_{\boldsymbol{z}_{*}|\mathcal{D}^{[i]}(t)}$, $\check{\sigma}'^2_{\boldsymbol{z}_{*}|\mathcal{D}^{[i]}(t)}=\check{\sigma}^2_{\boldsymbol{z}_{*}|\mathcal{D}^{[i]}(t)}$ where
\begin{align}\label{eq:5}
&\check{{\mu}}_{\boldsymbol{z}_{*}|\mathcal{D}^{[i]}(t)}\triangleq {{\rm ker}}(\boldsymbol{z}_{*},\boldsymbol{z}_*^{[i]}(t))\mathring{k}(\boldsymbol{z}_*^{[i]}(t),\boldsymbol{z}_*^{[i]}(t))^{-1}{y}_{\boldsymbol{z}_*^{[i]}(t)}^{[i]}, \nonumber \\
&\check{\sigma}^2_{\boldsymbol{z}_{*}|\mathcal{D}^{[i]}(t)}\triangleq {{\rm ker}}(\boldsymbol{z}_{*},\boldsymbol{z}_{*}) \nonumber \\
& -{{\rm ker}}(\boldsymbol{z}_*,\boldsymbol{z}_*^{[i]}(t))\mathring{k}(\boldsymbol{z}_*^{[i]}(t),\boldsymbol{z}_*^{[i]}(t))^{-1}{{\rm ker}}(\boldsymbol{z}_*^{[i]}(t),\boldsymbol{z}_{*})
\end{align}
with $\mathring{k}(\boldsymbol{z}_*^{[i]}(t),\boldsymbol{z}_*^{[i]}(t))\triangleq {{\rm ker}}(\boldsymbol{z}_*^{[i]}(t),\boldsymbol{z}_*^{[i]}(t))+(\sigma_e^{[i]})^2$. The predictive mean and variance of the Byzantine agents are written as $\check{{\mu}}'_{\boldsymbol{z}_{*}|\mathcal{D}^{[i]}(t)}=\star$ and $\check{\sigma}'^2_{\boldsymbol{z}_{*}|\mathcal{D}^{[i]}(t)}=\star$. The implementation of the agent-based local GPR is provided in Algorithm \ref{Alg:2}.

\subsubsection{Cloud-based aggregated GPR}
We develop a Byzantine-resilient PoE aggregation rule for the cloud. The algorithm is described as follows:

\emph{Step 1}: For each $\boldsymbol{z}_*\in\mathcal{Z}_{*}$, the cloud collects potentially corrupted mean $\check{{\mu}}'_{\boldsymbol{z}_{*}|\mathcal{D}^{[i]}(t)}$ and variance $\check{{\sigma}}'^2_{\boldsymbol{z}_{*}|\mathcal{D}^{[i]}(t)}$ from each agent $i$. 

\emph{Step 2}: The cloud constructs two agent sets to tolerate Byzantine attacks by removing the largest and smallest $\beta$ fraction of the local predictive means $\check{{\mu}}'_{\boldsymbol{z}_{*}|\mathcal{D}^{[i]}(t)}$ and variances $\check{{\sigma}}'^2_{\boldsymbol{z}_{*}|\mathcal{D}^{[i]}(t)}$, respectively. For $i\in\mathcal{V}$, we sort $\check{{\mu}}'_{\boldsymbol{z}_{*}|\mathcal{D}^{[i]}(t)}$ and $\check{{\sigma}}'^2_{\boldsymbol{z}_{*}|\mathcal{D}^{[i]}(t)}$ in non-descending order. We denote by $\mathcal{T}_{\max}^{\mu}(t)$ a set of $\beta n$ agent sets with the largest local predictive means, and by $\mathcal{T}_{\min}^{\mu}(t)$ a set of $\beta n$ agents with the smallest local predictive means. Similarly, we denote trimmed agent sets of variance by $\mathcal{T}_{\max}^{\sigma}(t)$ and $\mathcal{T}_{\min}^{\sigma}(t)$. 
Then, we define
\begin{align}\label{eq:6}
\mathcal{I}^{\mu}(t)&\triangleq\mathcal{V}\setminus(\mathcal{T}_{\max}^{\mu}(t)\bigcup\mathcal{T}_{\min}^{\mu}(t)),\nonumber \\ \mathcal{I}^{\sigma}(t)&\triangleq\mathcal{V}\setminus(\mathcal{T}_{\max}^{\sigma}(t)\bigcup\mathcal{T}_{\min}^{\sigma}(t)).
\end{align}

\emph{Step 3:} Define a global training dataset as $\mathcal{D}(t)\triangleq\bigcup_{i\in\mathcal{V}}\mathcal{D}^{[i]}(t)$, and a common set as  $\mathcal{I}(t)\triangleq\mathcal{I}^{\mu}(t)\bigcap\mathcal{I}^{\sigma}(t)$. 

Given $\mathcal{I}(t)$ and a test input $\boldsymbol{z}_*\in\boldsymbol{\mathcal{Z}}$, we propose a Byzantine-resilient PoE aggregation rule below
\begin{align}\label{eq:7}
\hat{{\mu}}_{\boldsymbol{z}_*|\mathcal{D}(t)}&\triangleq \frac{\hat{\sigma}^2_{\boldsymbol{z}_*|\mathcal{D}(t)}}{|\mathcal{I}(t)|}\sum_{i\in\mathcal{I}(t)}\check{{\mu}}'_{\boldsymbol{z}_{*}|\mathcal{D}^{[i]}(t)}\check{\sigma}'^{-2}_{\boldsymbol{z}_{*}|\mathcal{D}^{[i]}(t)},\nonumber \\
\hat{\sigma}^2_{\boldsymbol{z}_*|\mathcal{D}(t)}&\triangleq\frac{|\mathcal{I}(t)|}{\sum_{i\in\mathcal{I}(t)}\check{\sigma}'^{-2}_{\boldsymbol{z}_{*}|\mathcal{D}^{[i]}(t)}}.
\end{align}
Note that if all the agents are known to be benign, we can simply select $\beta=\alpha=0$, and hence $\mathcal{I}^{\mu}(t)=\mathcal{I}^{\sigma}(t)=\mathcal{V}$. In this case, the Byzantine-resilient PoE (\ref{eq:7}) reduces to the standard PoE in \cite{Hinton2002}. Then the cloud-based aggregated GPR returns $\hat{{\boldsymbol{\mu}}}_{\mathcal{Z}_{*}|\mathcal{D}(t)}\triangleq[\hat{{\mu}}_{\boldsymbol{z}_{*}|\mathcal{D}(t)}]_{\boldsymbol{z}_{*}\in\mathcal{Z}_{*}}$ and $\hat{{\boldsymbol{\sigma}}}^{2}_{\mathcal{Z}_{*}|\mathcal{D}(t)}\triangleq[\hat{{\sigma}}^2_{\boldsymbol{z}_{*}|\mathcal{D}(t)}]_{\boldsymbol{z}_{*}\in\mathcal{Z}_{*}}$. The implementation of the cloud-based aggregated GPR is provided in Algorithm \ref{Alg:3}.

\begin{algorithm}[!ht]
	\caption{Cloud-based aggregated GPR }
	\label{Alg:3}
	\begin{algorithmic}[1]
		\State cGPR$(\check{\boldsymbol{\mu}}'_{\mathcal{Z}_*|\mathcal{D}^{[i]}(t)},\check{\boldsymbol{\sigma}}'^{2}_{\mathcal{Z}_*|\mathcal{D}^{[i]}(t)})$
		
		\State $\mathcal{D}(t)=\bigcup_{i\in\mathcal{V}}\mathcal{D}^{[i]}(t)$
		\For{$\boldsymbol{z}_*\in\mathcal{Z}_*$}

		\State Construct sets $\mathcal{I}^{\mu}(t)$ and $\mathcal{I}^{\sigma}(t)$ according to (\ref{eq:6})
		
		\State Compute $\mathcal{I}(t)=\mathcal{I}^{\mu}(t)\bigcap\mathcal{I}^{\sigma}(t)$
		
		\State Implement aggregation rule (\ref{eq:7}) 
		
		\State Broadcast $\hat{{\boldsymbol{\mu}}}_{\mathcal{Z}_{*}|\mathcal{D}(t)}$ and $\hat{{\boldsymbol{\sigma}}}^2_{\mathcal{Z}_{*}|\mathcal{D}(t)}$  to all the agents
		
		\EndFor
		\State \Return $\hat{{\boldsymbol{\mu}}}_{\mathcal{Z}_{*}|\mathcal{D}(t)}$, $\hat{{\boldsymbol{\sigma}}}^2_{\mathcal{Z}_{*}|\mathcal{D}(t)}$
	\end{algorithmic}
\end{algorithm}


The following {{assumption and}} lemma characterize {{$|\mathcal{I}(t)|$}}.

\begin{Asu}\label{Asu:00}
	It holds that $0<\alpha\leq\beta<\frac{1}{4}$.
\end{Asu}

The assumption $\beta<\frac{1}{4}$ in Assumption \ref{Asu:00} ensures that $(1-4\beta)n>0$ and $|\mathcal{I}(t)|\neq 0$.

\begin{Lem}\label{Lem:22}
	Suppose that $\beta<\frac{1}{4}$, it holds that $n-4\beta n\le|\mathcal{I}(t)|\le n-2\beta n$ at each time instant $t$.
\end{Lem}

The proof of Lemma \ref{Lem:22} is presented in Section 5.1. 

\begin{Rem}[Distinction between classic Byzantine-resilient algorithms]
Byzantine-resilient aggregation rules in the existing works \cite{SPLM2022}, \cite{ESP2022}, \cite{allouah2023fixing} and \cite{LLXW2019} refer to element-wise gradient computations. Note that these elements are decoupled, and the aggregation of one element for all agents does not depend on other elements.

In contrast, Byzantine-resilient PoE aggregation in (\ref{eq:7}) is coupled by taking a weighted average of the local predictive means $\check{{\mu}}'_{\boldsymbol{z}_{*}|\mathcal{D}^{[i]}(t)}$ where the weights are the local predictive variances $\check{\sigma}'^{2}_{\boldsymbol{z}_{*}|\mathcal{D}^{[i]}(t)}$. 

On the other side, most existing works on Byzantine-resilient algorithms, e.g.,   \citep{BEG2017, YLJ2017, YCRB2018}, can tolerate Byzantine agents with $\alpha<\frac{1}{2}$. The variables in \cite{BEG2017}, \cite{YLJ2017}, \cite{YCRB2018} refer to the elements of a multi-dimensional gradient. Therefore, the decoupling element of the gradient renders that the Byzantine-resilient algorithms in \cite{BEG2017}, \cite{YLJ2017}, \cite{YCRB2018} only need to remove $2\alpha n$ extreme values. If $\alpha<\frac{1}{2}$, there exists at least one agent whose element of the gradient is upper bounded by the values of the benign agents. 

In contrast, the Byzantine-resilient PoE aggregation (\ref{eq:7}) couples the local predictive means $\check{{\mu}}'_{\boldsymbol{z}_{*}|\mathcal{D}^{[i]}(t)}$ and the local predictive variances $\check{\sigma}'^{2}_{\boldsymbol{z}_{*}|\mathcal{D}^{[i]}(t)}$. After removing $2\alpha n$ maximal and minimal predictive means with $\alpha<\frac{1}{4}$, the set $\mathcal{I}^{\mu}(t)$ in (\ref{eq:6}) includes more than one half agents whose predictive means are upper bounded by those of the benign agents. Likewise, after removing $2\alpha n$ maximal and minimal predictive variances with $\alpha<\frac{1}{4}$, the set $\mathcal{I}^{\sigma}(t)$ in (\ref{eq:6}) includes more than one half agents whose predictive variances are upper bounded by those of the benign agents. With $\alpha<\frac{1}{4}$, the elimination of $4\alpha n$ extreme values ensures that the common set $\mathcal{I}(t)$ in step 3 has at least one agent whose both local predictive means $\check{{\mu}}'_{\boldsymbol{z}_{*}|\mathcal{D}^{[i]}(t)}$ and variances $\check{\sigma}'^{2}_{\boldsymbol{z}_{*}|\mathcal{D}^{[i]}(t)}$ are upper bounded by the values of the benign agents, and thus used for the Byzantine-resilient PoE aggregation.
\end{Rem}

Lemma \ref{Lem:0} shows that $\check{{\mu}}'_{\boldsymbol{z}_{*}|\mathcal{D}^{[i]}(t)}$ and $\check{{\sigma}}'^2_{\boldsymbol{z}_{*}|\mathcal{D}^{[i]}(t)}$ for $i\in\mathcal{I}(t)$ are bounded by the values of the benign agents.

\begin{Lem}\label{Lem:0}
	Let Assumption \ref{Asu:00} hold. For each $i\in\mathcal{I}(t)$, we have $\min_{j\in\mathcal{V}\backslash\mathcal{B}}\left\{\check{{\mu}}_{\boldsymbol{z}_{*}|\mathcal{D}^{[j]}(t)}\right\}\le\check{{\mu}}'_{\boldsymbol{z}_{*}|\mathcal{D}^{[i]}(t)}\le\max_{j\in\mathcal{V}\backslash\mathcal{B}}\left\{\check{{\mu}}_{\boldsymbol{z}_{*}|\mathcal{D}^{[j]}(t)}\right\}$ and  $\min_{j\in\mathcal{V}\backslash\mathcal{B}}\left\{\check{{\sigma}}^2_{\boldsymbol{z}_{*}|\mathcal{D}^{[j]}(t)}\right\}\le\check{{\sigma}}'^2_{\boldsymbol{z}_{*}|\mathcal{D}^{[i]}(t)}\le\max_{j\in\mathcal{V}\backslash\mathcal{B}}\left\{\check{{\sigma}}^2_{\boldsymbol{z}_{*}|\mathcal{D}^{[j]}(t)}\right\}$.
\end{Lem}

The proof of Lemma \ref{Lem:0} is presented in Section 5.2.

\subsubsection{Agent-based fused GPR}
Each agent $i$ refines its predictions through fusing those of the agent-based local GPR and the cloud-based aggregated GPR. More specifically, agent $i$ needs to construct $\mathcal{Z}_{fused}^{[i]}(t)$, a set of inputs where the cloud-based aggregated GPR performs better. {{Theorem \ref{Thm:3} in Section 4.2 provides different guarantees for different construction of $\mathcal{Z}_{fused}^{[i]}(t)$.}} If $\boldsymbol{z}_*\in\mathcal{Z}_{fused}^{[i]}(t)$, the predictions from the cloud-based aggregated GPR are used. Otherwise, those from the agent-based local GPR are used. The refined predictions in the agent-based fused GPR will not be transmitted back to the cloud, hence they cannot change the final aggregation. The implementation of the agent-based fused GPR is provided in Algorithm \ref{Alg:4}.

\begin{algorithm}[!ht]
	\caption{Agent-based fused GPR }
	\label{Alg:4}
	\begin{algorithmic}[1]
		\State$\text{fGPR}(\check{\boldsymbol{\mu}}_{\mathcal{Z}_*|\mathcal{D}^{[i]}(t)},\check{\boldsymbol{\sigma}}^2_{\mathcal{Z}_*|\mathcal{D}^{[i]}(t)},
		\hat{\boldsymbol{\mu}}_{\mathcal{Z}_{*}|\mathcal{D}(t)},
		\hat{\boldsymbol{\sigma}}^2_{\mathcal{Z}_{*}|\mathcal{D}(t)})
		$
		
		\State Construct $\mathcal{Z}_{fused}^{[i]}(t)$
		\For{$\boldsymbol{z}_*\in\mathcal{Z}_*$}
		\If{$\boldsymbol{z}_*\in\mathcal{Z}_{fused}^{[i]}(t)$}
		\State Compute
		\begin{align*}
			\tilde{{\mu}}^{[i]}_{\boldsymbol{z}_*|\mathcal{D}(t)}, (\tilde{\sigma}^{[i]}_{\boldsymbol{z}_*|\mathcal{D}(t)})^{2}&\leftarrow\left\{\begin{array}{l}
			\underbrace{\hat{\mu}_{\boldsymbol{z}_{*}|\mathcal{D}(t)}, \hat{\sigma}^2_{\boldsymbol{z}_{*}|\mathcal{D}(t)}}_\text{Benign agent},\\
			\star~~~~~~~~~ \text{Byzantine agent}
			\end{array}\right. 
		\end{align*}
		\Else
		
		\State Compute
		\begin{align*}
			\tilde{{\mu}}^{[i]}_{\boldsymbol{z}_*|\mathcal{D}(t)},(\tilde{\sigma}^{[i]}_{\boldsymbol{z}_*|\mathcal{D}(t)})^{2}&\leftarrow\left\{\begin{array}{l}
			\underbrace{\check{\mu}_{\boldsymbol{z}_*|\mathcal{D}^{[i]}(t)},\check{\sigma}^2_{\boldsymbol{z}_{*}|\mathcal{D}^{[i]}(t)}}_\text{Benign agent},\\
			\star~~~~~~~~~~ \text{Byzantine agent}
			\end{array}\right.
		\end{align*}
		\EndIf
		\EndFor
	
		\State \Return $\tilde{\boldsymbol{\mu}}^{[i]}_{\mathcal{Z}_*|\mathcal{D}(t)}$, $(\tilde{\boldsymbol{\sigma}}^{[i]}_{\mathcal{Z}_*|\mathcal{D}(t)})^2$
	\end{algorithmic}
\end{algorithm}
\subsection{Performance guarantee}
This section discusses the theoretical results for the robustness of the cloud-based aggregated GPR and the learning performance improvements of the agent-based fused GPR when a fraction of agents are Byzantine agents. To increase readability, Table \ref{Tab:0} lists the symbols that are used in multiple important results.

\begin{table*}[!t]
	\centering\caption{Notation used in theory} 
	\label{Tab:0}
	\centering
	\begin{tabular}{|l|l|}
		\hline
		Symbol  & Meaning  \\
		\hline
        $\alpha$ & Ratio of the number of Byzantine agents to the total number of agents \\
		\hline
	$\beta$	& Fraction of the largest and smallest predictions to be removed\\
		\hline
		$\gamma_d$ &  Upper bound of dispersion\\
		\hline
        $\check{{\mu}}'_{\boldsymbol{z}_{*}|\mathcal{D}^{[i]}(t)}\backslash\check{{\mu}}_{\boldsymbol{z}_{*}|\mathcal{D}^{[i]}(t)}$ & Attacked$\backslash$attack-free predictive mean from the agent-based local GPR \\
		\hline
		$\hat{\mu}_{\boldsymbol{z}_*|\mathcal{D}(t)}$ & Predictive mean from the cloud-based aggregated GPR \\
		\hline
		$\tilde{{\mu}}_{\boldsymbol{z}_*|\mathcal{D}^{[i]}(t)}$ &  Predictive mean from the agent-based fused GPR \\
		\hline
		$\check{{\sigma}}'^2_{\boldsymbol{z}_{*}|\mathcal{D}^{[i]}(t)}\backslash \check{{\sigma}}^2_{\boldsymbol{z}_{*}|\mathcal{D}^{[i]}(t)}$ & Attacked$\backslash$attack-free predictive variance from the agent-based local GPR  \\
		\hline
		$\hat{\sigma}^2_{\boldsymbol{z}_*|\mathcal{D}(t)}$ & Predictive variance from the cloud-based aggregated GPR \\
		\hline
		$(\tilde{\sigma}^{[i]}_{\boldsymbol{z}_*|\mathcal{D}(t)})^{2}$ & Predictive variance from the agent-based fused GPR \\
		\hline
		$(\sigma_e^{[i]})^2$ &  Variance of the noise \\
		\hline
		$(\sigma_e^{\max})^2\backslash(\sigma_e^{\min})^2$ &  Maximal$\backslash$minimal variance of the noise over all benign agents \\
		\hline
		$d^{[i]}(t)\backslash d^{\max}(t)$ & Dispersion of training dataset $\mathcal{D}^{[i]}(t)$ $\backslash$ Maximal dispersion over all training datasets  \\
		\hline
		$\mathcal{D}^{[i]}(t)$ & Training dataset of agent $i$\\
		\hline 
		$\mathcal{D}(t)$ & A union of training datasets of all agents \\
		\hline
		$\mathcal{I}^{\mu}(t)\backslash\mathcal{I}^{\sigma}(t)$ & Agent set with $2\beta n$ extreme local predictive means$\backslash$variances removed \\
		\hline
		$\mathcal{I}(t)$ & Agent set with $2\beta n$ extreme local predictive means and variances removed \\
		\hline
		$\ell_{\eta}$ &  Lipschitz constant  \\
		\hline
		$\mathcal{T}_{\max}^{\mu}(t)\backslash\mathcal{T}_{\min}^{\mu}(t)$ & Agent set with the largest$\backslash$smallest $\beta n$ local predictive means \\
		\hline
		$\mathcal{T}_{\max}^{\sigma}(t)\backslash\mathcal{T}_{\min}^{\sigma}(t)$ & Agent set with the largest$\backslash$smallest $\beta n$ local predictive variances \\
		\hline
	\end{tabular}
\end{table*}

We start with the assumptions of the kernel function ${{\rm ker}}$.
		\begin{Asu}\label{Asu:2}
			1) (Decomposition). The kernel function ${{\rm ker}}(\cdot,\cdot)$ can be decomposed in such a way that ${{\rm ker}}(\cdot,\cdot)=\kappa(D(\cdot,\cdot))$, where $\kappa:\mathbb{R}_{+}\rightarrow\mathbb{R}_{++}$ is continuous.
			
			2) (Monotonicity). It holds that $\kappa(s)$ is monotonically decreasing in $s$ and $\kappa(0)=\sigma_f^2$.
		\end{Asu}
		
		Many existing kernels satisfy Assumption \ref{Asu:2}, e.g., squared exponential kernel $
		k(\boldsymbol{z},\boldsymbol{z}_*) = \sigma_f^2\exp({-\frac{1}{2\ell^2}}(\boldsymbol{z}-\boldsymbol{z}_*)^2)$ \citep{CC2006}. We
		use the following assumption in the analysis.
		\begin{Asu}\label{Asu:3}
			It satisfies that $\eta\sim\mathcal{G}\mathcal{P}(0,{{\rm ker}})$ and $\eta$ is Lipschitz continuous on $\boldsymbol{\mathcal{Z}}$.
		\end{Asu}
		
The target function $\eta$ is completely specified by Gaussian process with kernel function ${{\rm ker}}$. This assumption is common in the analysis of GPR \citep{NASM2012}. The Lipschitz continuity of $\eta$ implies that there exists some positive constant $\ell_{\eta}$ such that $|\eta(\boldsymbol{z})-\eta(\boldsymbol{z}')|\le\ell_{\eta}D(\boldsymbol{z},\boldsymbol{z}')$, $\forall \boldsymbol{z},\boldsymbol{z}'\in\mathcal{Z}$. For $i\in\mathcal{V}$, we define the dispersion of local data as $ d^{[i]}(t)\triangleq \sup_{\boldsymbol{z}\in\boldsymbol{\mathcal{Z}}}D(\boldsymbol{z},\mathcal{Z}^{[i]}(t))$, which can be interpreted as a measurement of how dense the sampled
data are distributed within a compact space. We denote the maximal local dispersion as $d^{\max}(t)\triangleq\max_{i\in\mathcal{V}}\{d^{[i]}(t)\}$ and the parameter as $\sigma\triangleq\frac{\sigma_f^2\sigma_e^{\max}}{\sigma_f^2+(\sigma_e^{\min})^2}$ where{{ $\sigma_e^{\max}\triangleq\max_{i\in\mathcal{V}\setminus\mathcal{B}}\{\sigma_e^{[i]}\}$ and $\sigma_e^{\min}\triangleq\min_{i\in\mathcal{V}\setminus\mathcal{B}}\{\sigma_e^{[i]}\}$.}} Theorem \ref{Thm:2} derives bounds on the prediction error and the global predictive variance.

\begin{Thm}\label{Thm:2}
	\textbf{Part I:}(Cloud-based aggregated GPR: Mean) Suppose Assumptions \ref{Asu:00}, \ref{Asu:2} and \ref{Asu:3} hold. For any $\boldsymbol{z}_*\in\boldsymbol{\mathcal{Z}}$ and $0<\delta<1$, with probability at least $1-\delta$, it holds that $\left|\hat{{\mu}}_{\boldsymbol{z}_*|\mathcal{D}(t)}-\eta(\boldsymbol{z}_*)\right|\le\Theta(d^{\max}(t))+\Delta(d^{\max}(t))
	$ where $\Theta(s)\triangleq(1-\frac{\kappa(s)}{\sigma_f^2+(\sigma_e^{\max})^2})\|\eta\|_{\infty}+\frac{\sigma_f^2\ell_{\eta} s}{\sigma_f^2+(\sigma_e^{\min})^2}+\sqrt{2\sigma^2(\ln2-\ln\delta)}$ and $\Delta(s)\triangleq\frac{\sigma_f^4+\sigma_f^2(\sigma_e^{\max})^2-\kappa(s)^2}{\sigma_f^2(\sigma_e^{\min})^2}\frac{2\alpha}{1-4\beta}\notag\\{{\Theta(s)}}$.
	
	\textbf{Part II:} (Cloud-based aggregated GPR: Variance) Let Assumptions \ref{Asu:00} and \ref{Asu:2} hold. For any $\boldsymbol{z}_*\in\boldsymbol{\mathcal{Z}}$, it holds that ${{(1-2\beta)}}\left(\frac{\sigma_f^2(\sigma_e^{\min})^2}{\sigma_f^2+(\sigma_e^{\max})^2}\right)\le\hat{\sigma}^2_{\boldsymbol{z}_*|\mathcal{D}(t)}\le {{\frac{1}{1-4\beta}}}\left(\sigma_f^2-\frac{\kappa(d^{\max}(t))^2}{\sigma_f^2+(\sigma_e^{\max})^2}\right)$.
\end{Thm}

 The proof of Theorem \ref{Thm:2} is presented in Section 5.3. In part I, the term $\Delta(d^{\max}(t))$ characterizes the performance loss caused by Byzantine agents. It is zero for the attack-free case, i.e., $\alpha=\beta=0$. For this case,
%
%
the upper bound of Part I becomes $\Theta(d^{\max}(t))$. We assume $\lim_{t\rightarrow\infty}d^{[i]}(t)=0$ for $i\in\mathcal{V}$, the limit of the upper bound is $\frac{(\sigma_e^{\max})^2}{\sigma_f^2+(\sigma_e^{\max})^2}\|\eta\|_{\infty}+\sqrt{2\sigma^2(\ln2-\ln\delta)}$. Recall that $\sigma=\frac{\sigma_f^2\sigma_e^{\max}}{\sigma_f^2+(\sigma_e^{\min})^2}$. Then we can see that the limit is positively related to the maximal variance $(\sigma_e^{\max})^2$ of the noise. 
Notice that $\Delta(d^{\max}(t))$ is positively related to $\alpha$, which indicates a trade-off between the prediction error and the number of Byzantine agents. With a larger $\alpha$, our algorithm can tolerate more Byzantine agents, but the prediction error gets larger. On the other hand, the prediction performance improves if the number of Byzantine agents is decreasing. 
In addition, when $\alpha$ is fixed, $\Delta(d^{\max}(t))$ is proportional to $\beta$. This implies that learning performance improves if the set $\mathcal{I}(t)$ expands.

The upper and lower bounds in Part II {{depend on}} $\beta$. {{With a larger $\beta$, the prediction uncertainty characterized by $\hat{\sigma}^2_{\boldsymbol{z}_*|\mathcal{D}(t)}$ is higher in that more predictive information are removed by the Byzantine-resilient PoE. As $\beta$ decreases, the lower and upper bounds will approach those in the attack-free case}}. As $t$ goes to infinity, we have $\limsup_{t\rightarrow\infty}\hat{\sigma}^2_{\boldsymbol{z}_*|\mathcal{D}(t)}\le\frac{\sigma_f^2(\sigma_e^{\max})^2}{\sigma_f^2+(\sigma_e^{\max})^2}$. This implies that the prediction uncertainty diminishes as the variance of measurement noise decreases. {{Recall that $\sigma_e^{\max}=\max_{i\in\mathcal{V}\setminus\mathcal{B}}\{\sigma_e^{[i]}\}$ and $\sigma_e^{\min}=\min_{i\in\mathcal{V}\setminus\mathcal{B}}\{\sigma_e^{[i]}\}$}}. If only the variance $(\sigma_e^{[i]})^2$ of the noise is perturbed, the bounds show that the performance will not impacted.

It is worth noting that {{the cloud-based aggregated GPR makes predictions using all the data while the agent-based local GPR makes predictions using local data only. Therefore, the cloud-based aggregated GPR tends to have better learning performances compared with the agent-based local GPR  \citep{HJYY2018}. Since eventually it is the agents who make predictions, the federation of the agents allows the agents to utilize the predictions from the cloud to enhance their local prediction accuracy by leveraging the agent-based fused GPR.}}


Next, we discuss how the {{fusion algorithms}} of the agent-based fused GPR for the benign agents improve the learning performances. Since $\mathcal{Z}$ is compact, $d^{[i]}(t)$ is uniformly bounded and the upper bound is denoted by $\gamma_d$. We give two possible ways to construct
\begin{align}
&\mathcal{Z}_{fused}^{[i]}(t) \triangleq \{\boldsymbol{z}_*\in\mathcal{Z}_*|\check{\sigma}^2_{\boldsymbol{z}_*|\mathcal{D}^{[i]}(t)}>\hat{\sigma}^2_{\boldsymbol{z}_{*}|\mathcal{D}(t)}\}  \label{eq:88}\\
&\mathcal{Z}_{fused}^{[i]}(t) \nonumber \\
&\quad\triangleq \{\boldsymbol{z}_*\in\mathcal{Z}_*|\hat{\theta}\hat{\sigma}^2_{\boldsymbol{z}_{*}|\mathcal{D}(t)}-\check{\theta}^{[i]}(t)\check{\sigma}^2_{\boldsymbol{z}_*|\mathcal{D}^{[i]}(t)}<-\Gamma\} \label{eq:99}
\end{align} 
where $\Gamma \triangleq \frac{\sigma_f^2\ell_{\eta} \gamma_d}{\sigma_f^2+(\sigma_e^{\min})^2}+\sqrt{2\sigma^2(\ln2-\ln\delta)}$ with $0<\delta<1$, $\hat{\theta}\triangleq\frac{\sigma_f^2+(\sigma_e^{\max})^2-\kappa(\gamma_d)}{{{(1-2\beta)}}\sigma_f^2(\sigma_e^{\min})^2}\|\eta\|_{\infty}$, $\check{\theta}^{[i]}(t)\triangleq\frac{\sigma_e^{[i]}\kappa(d^{[i]}(t))}{\sigma_f^4+\sigma_f^2(\sigma_e^{[i]})^2-\kappa(d^{[i]}(t))^2}$.  

Part I of Theorem \ref{Thm:3} first provides an upper bound of the prediction error from the agent-based fused GPR by using (\ref{eq:88}). Part II of Theorem \ref{Thm:3} constructs (\ref{eq:99}), and quantifies the improvements of the learning performance of the agent-based fused GPR by {{proving}} the mean squared errors $\mathbb{E}[(\hat{\mu}_{\boldsymbol{z}_{*}|\mathcal{D}(t)}-\eta(\boldsymbol{z}_*))^2]$ {{is smaller than}} $\mathbb{E}[(\check{{\mu}}_{\boldsymbol{z}_*|\mathcal{D}^{[i]}(t)}-\eta(\boldsymbol{z}_*))^2]$. Notice that the two parts are complementary. Part II quantifies the learning performance improvements when Lipschitz constant $\ell_{\eta}$ is known. Part III of Theorem \ref{Thm:3} derives bounds on the predictive variance of the agent-based fused GPR.

\begin{Thm}\label{Thm:3}

	\textbf{Part I:} (Agent-based fused GPR: Mean) Suppose Assumptions \ref{Asu:00}, \ref{Asu:2}, \ref{Asu:3} hold and let $\mathcal{Z}_{fused}^{[i]}(t)=\{\boldsymbol{z}_*\in\mathcal{Z}_*|\check{\sigma}^2_{\boldsymbol{z}_*|\mathcal{D}^{[i]}(t)}>\hat{\sigma}^2_{\boldsymbol{z}_{*}|\mathcal{D}(t)}\}$. For all $\boldsymbol{z}_*\in\boldsymbol{\mathcal{Z}}$, $0<\delta<1$ and $i\in\mathcal{V}\backslash\mathcal{B}$, with probability at least $1-\delta$, it holds that
	\begin{itemize}
		\item [] $|\tilde{{\mu}}^{[i]}_{\boldsymbol{z}_*|\mathcal{D}(t)}-\eta(\boldsymbol{z}_*)|\le\Theta(d^{\max}(t))+\Delta(d^{\max}(t))$.
	\end{itemize}
	\textbf{Part II:} (Agent-based fused GPR: Mean) Suppose Assumptions \ref{Asu:00}, \ref{Asu:2}, \ref{Asu:3} hold and let $\mathcal{Z}_{fused}^{[i]}(t)=\{\boldsymbol{z}_*\in\mathcal{Z}_*|\hat{\theta}\hat{\sigma}^2_{\boldsymbol{z}_{*}|\mathcal{D}(t)}-\check{\theta}^{[i]}(t)\check{\sigma}^2_{\boldsymbol{z}_*|\mathcal{D}^{[i]}(t)}<-\Gamma\}$. For any $i\in\mathcal{V}\backslash\mathcal{B}$ and $0<\delta<1$, if $\boldsymbol{z}_*\in\mathcal{Z}_{fused}^{[i]}(t)$, the following holds with probability at least $1-\delta$
	$$\mathbb{E}[(\tilde{\mu}^{[i]}_{\boldsymbol{z}_{*}|\mathcal{D}(t)}-\eta(\boldsymbol{z}_*))^2]-\mathbb{E}[(\check{{\mu}}_{\boldsymbol{z}_*|\mathcal{D}^{[i]}(t)}-\eta(\boldsymbol{z}_*))^2]<0;$$ otherwise,
	$\mathbb{E}[(\tilde{\mu}^{[i]}_{\boldsymbol{z}_{*}|\mathcal{D}(t)}-\eta(\boldsymbol{z}_*))^2]=\mathbb{E}[(\check{{\mu}}_{\boldsymbol{z}_*|\mathcal{D}^{[i]}(t)}-\eta(\boldsymbol{z}_*))^2].$

	\textbf{Part III:} (Agent-based fused GPR: Variance) Let Assumptions \ref{Asu:00} and \ref{Asu:2} hold and consider both $\mathcal{Z}_{fused}^{[i]}(t)$ in Parts I and II. For all $\boldsymbol{z}_*\in\boldsymbol{\mathcal{Z}}$ and $i\in\mathcal{V}\backslash\mathcal{B}$, it holds that ${{(1-2\beta)}}\left(\frac{\sigma_f^2(\sigma_e^{\min})^2}{\sigma_f^2+(\sigma_e^{\max})^2}\right)\le(\tilde{\sigma}^{[i]}_{\boldsymbol{z}_*|\mathcal{D}(t)})^{2}\le {{\frac{1}{1-4\beta}}}\left(\sigma_f^2-\frac{\kappa(d^{\max}(t))^2}{\sigma_f^2+(\sigma_e^{\max})^2}\right)$.
\end{Thm} 

The proof of Theorem \ref{Thm:3} is presented in Section 5.4. 

{{
\begin{Rem}
	All proofs of Theorem \ref{Thm:2} and Theorem \ref{Thm:3} hold only for the single-dimensional GPR. The extension to multi-dimensional GPRs is straightforward, since we can perform {{the proposed algorithm}} for each element of the latent function. It is a typical practice in literature {{(see \cite{HMF2004}, \cite{HJYY2018}, \cite{nguyen2009model}, \cite{NASM2012}, \cite{sui2015safe} and references therein)}} to study the single-dimension GPR in order to simplify notations. 
\end{Rem}

}}

\section{Proofs}
This section provides complete proofs for each lemma and theorem.

\subsection{Proof of Lemma 1}
Notice that  $|\mathcal{I}(t)|=|\mathcal{I}^{\mu}(t)\bigcap\mathcal{I}^{\sigma}(t)|\le|\mathcal{I}^{\mu}(t)|$. By definition of $\mathcal{I}^{\mu}(t)(t)$ in (\ref{eq:6}), we have $\mathcal{I}^{\mu}(t)=\mathcal{V}\setminus(\mathcal{T}^{\mu}_{\max}(t)\bigcup\mathcal{T}^{\mu}_{\min}(t))$. Definitions of $\mathcal{T}^{\mu}_{\max}(t)$ and $\mathcal{T}^{\mu}_{\min}(t)$ give $|\mathcal{T}^{\mu}_{\max}(t)|=|\mathcal{T}^{\mu}_{\min}(t)|=\beta n$. Then we have $|\mathcal{I}^{\mu}(t)|=|\mathcal{V}\setminus(\mathcal{T}^{\mu}_{\max}(t)\bigcup\mathcal{T}^{\mu}_{\min}(t))|= n-2\beta n$, which indicates that $|\mathcal{I}(t)|\le n-2\beta n$. 

Since $\mathcal{I}^{\mu}(t)\bigcup\mathcal{I}^{\sigma}(t)\subseteq\mathcal{V}$, then it holds $|\mathcal{I}^{\mu}(t)\bigcup\mathcal{I}^{\sigma}(t)|\le n$. By definition of $\mathcal{I}^{\sigma}(t)$ in (\ref{eq:6}), we have  $|\mathcal{I}^{\sigma}(t)|=n-2\beta n$. Therefore, we have $|\mathcal{I}(t)|=|\mathcal{I}^{\mu}(t)\bigcap\mathcal{I}^{\sigma}(t)|=|\mathcal{I}^{\mu}(t)|+|\mathcal{I}^{\sigma}(t)|-|\mathcal{I}^{\mu}(t)\bigcup\mathcal{I}^{\sigma}(t)|\ge n-4\beta n$. 

\subsection{Proof of Lemma 2}
Notice that $\mathcal{I}(t)=\mathcal{I}^{\mu}(t)\bigcap\mathcal{I}^{\sigma}(t)$. Recall that in Step 2 of Section {{4.1.2}}, $\check{{\mu}}'_{\boldsymbol{z}_{*}|\mathcal{D}^{[i]}(t)}$ is sorted in non-descending order. Without loss of generality, agent $1$ has the smallest value and agent $n$ has the largest value, i.e.,  $\check{{\mu}}'_{\boldsymbol{z}_{*}|\mathcal{D}^{[i]}(t)}\leq\check{{\mu}}'_{\boldsymbol{z}_{*}|\mathcal{D}^{[i+1]}(t)}$ for $i=1,2,\ldots,n-1$. {{We consider the following two cases.
		
\emph{Case 1: $\mathcal{T}^{\mu}_{\max}(t)$ includes benign agents.} We let agent $q\in\mathcal{T}^{\mu}_{\max}(t)$ be benign, such that $\check{{\mu}}'_{\boldsymbol{z}_{*}|\mathcal{D}^{[q]}(t)}=\check{{\mu}}_{\boldsymbol{z}_{*}|\mathcal{D}^{[q]}(t)}$. {{Recall that  $\mathcal{T}^{\mu}_{\max}(t)$ contains $\beta n$ agents with the largest local predictive means. For any $q\in\mathcal{T}^{\mu}_{\max}(t)$ and $q'\in\mathcal{I}^{\mu}(t)$, we have $\check{{\mu}}'_{\boldsymbol{z}_{*}|\mathcal{D}^{[q']}(t)}\leq\check{{\mu}}'_{\boldsymbol{z}_{*}|\mathcal{D}^{[q]}(t)}=\check{{\mu}}_{\boldsymbol{z}_{*}|\mathcal{D}^{[q]}(t)}\leq\max_{j\in\mathcal{V}\backslash\mathcal{B}}\left\{\check{{\mu}}_{\boldsymbol{z}_{*}|\mathcal{D}^{[j]}(t)}\right\}$}}, {{which is consistent with the lemma}}.

\emph{Case 2: $\mathcal{T}^{\mu}_{\max}(t)$ excludes benign agents.} Suppose that there exists $i\in\mathcal{I}^{\mu}(t)$ such that  $\check{{\mu}}'_{\boldsymbol{z}_{*}|\mathcal{D}^{[i]}(t)}>\max_{j\in\mathcal{V}\backslash\mathcal{B}}\left\{\check{{\mu}}_{\boldsymbol{z}_{*}|\mathcal{D}^{[j]}(t)}\right\}$, then we have $i\in\mathcal{B}$. For all $q\in\mathcal{T}_{\max}^{\mu}(t)$, we have $\check{{\mu}}'_{\boldsymbol{z}_{*}|\mathcal{D}^{[q]}(t)}\geq\check{{\mu}}'_{\boldsymbol{z}_{*}|\mathcal{D}^{[i]}(t)}>\max_{j\in\mathcal{V}\backslash\mathcal{B}}\left\{\check{{\mu}}_{\boldsymbol{z}_{*}|\mathcal{D}^{[j]}(t)}\right\}$. Since $|\mathcal{T}_{\max}^{\mu}(t)|=\beta n$, then we have $\alpha n\ge \beta n+1$. It contradicts with $\alpha\le\beta$ in Assumption \ref{Asu:00}. }}

Therefore, we have $\check{{\mu}}'_{\boldsymbol{z}_{*}|\mathcal{D}^{[i]}(t)}\le\max_{j\in\mathcal{V}\backslash\mathcal{B}}\left\{\check{{\mu}}_{\boldsymbol{z}_{*}|\mathcal{D}^{[j]}(t)}\right\}$ for all $i \in \mathcal{I}^{\mu}(t)$. Likewise, we have $\min_{j\in\mathcal{V}\backslash\mathcal{B}}\left\{\check{{\mu}}_{\boldsymbol{z}_{*}|\mathcal{D}^{[j]}(t)}\right\}\\\le\check{{\mu}}'_{\boldsymbol{z}_{*}|\mathcal{D}^{[i]}(t)}$ for all $i \in \mathcal{I}^{\mu}(t)$.

Analogously, we conclude that $\min_{j\in\mathcal{V}\backslash\mathcal{B}}\left\{\check{{\sigma}}^2_{\boldsymbol{z}_{*}|\mathcal{D}^{[j]}(t)}\right\}\le\check{{\sigma}}'^2_{\boldsymbol{z}_{*}|\mathcal{D}^{[i]}(t)}\le\max_{j\in\mathcal{V}\backslash\mathcal{B}}\left\{\check{{\sigma}}^2_{\boldsymbol{z}_{*}|\mathcal{D}^{[i]}(t)}\right\}$ for all $i\in\mathcal{I}^{\sigma}(t)$.

\subsection{Proof of Theorem 1}
\textbf{Part I:} Roadmap of the proof: We first show in Lemma \ref{Lem:3} that at time instant $t$, the local predictive mean of agent $i\in\mathcal{V}$ in the attack-free scenario is a sub-Gaussian random variable. Then notice that by {{triangle}} inequality, the prediction errors under attacks can be bounded by the magnitude of attacks plus the prediction errors in the attack-free case. Therefore, we derive the upper bound of the prediction error in the attack-free case in Lemma \ref{Lem:5}. {{Lemma \ref{Lem:55} and Lemma \ref{Lem:4} are used to}} quantify the upper bound of the Byzantine attacks. 

\begin{Lem}\label{Lem:3}
	Let Assumptions \ref{Asu:2} and \ref{Asu:3} hold. For agent $i\in\mathcal{V}$ and $\boldsymbol{z}_*\in{\mathcal{Z}}_*$, it holds that  $\check{{\mu}}_{\boldsymbol{z}_*|\mathcal{D}^{[i]}(t)}$ is a sub-Gaussian random variable.
\end{Lem}

\emph{Proof:} Pick any $i\in\mathcal{V}$. Monotonicity of Assumption \ref{Asu:2} implies that $k(\boldsymbol{z}_*^{[i]}(t),\boldsymbol{z}_*^{[i]}(t))=\kappa(0)=\sigma_f^2$. For notational simplicity, we denote the distance by $D_{\boldsymbol{z}_*}^{\mathcal{Z}^{[i]}(t)}\triangleq D(\boldsymbol{z}_*,\mathcal{Z}^{[i]}(t))$. Hence, by (\ref{eq:5}), the local predictive mean is computed as $\check{{\mu}}_{\boldsymbol{z}_{*}|\mathcal{D}^{[i]}(t)}=\frac{\kappa(D_{\boldsymbol{z}_*}^{\mathcal{Z}^{[i]}(t)})}{\sigma_f^2+(\sigma_e^{[i]})^2}{y}_{\boldsymbol{z}_*^{[i]}(t)}^{[i]}$. Given the observation model (\ref{eq:4}), ${y}_{\boldsymbol{z}_*^{[i]}(t)}^{[i]}$ can be decomposed into a deterministic process $\eta(\boldsymbol{z}_*^{[i]}(t))$ and a zero-mean Gaussian noise ${y}_{\boldsymbol{z}_*^{[i]}(t)}^{[i]}-\eta(\boldsymbol{z}_*^{[i]}(t))$. For agent $i\in\mathcal{V}$, we denote the expectation and variance of $\check{{\mu}}_{\boldsymbol{z}_*|\mathcal{D}^{[i]}(t)}$ by $\mathbb{E}\left[\check{{\mu}}_{\boldsymbol{z}_*|\mathcal{D}^{[i]}(t)}\right]$ and $Var\left[\check{{\mu}}_{\boldsymbol{z}_*|\mathcal{D}^{[i]}(t)}\right]$, respectively. Notice that, ${y}_{\boldsymbol{z}_*^{[i]}(t)}^{[i]}-\eta(\boldsymbol{z}_*^{[i]}(t))$ is the random variable with variance $(\sigma_e^{[i]})^2$, and this implies $\check{{\mu}}_{\boldsymbol{z}_{*}|\mathcal{D}^{[i]}(t)}\sim\mathcal{N}\left(\frac{\kappa(D_{\boldsymbol{z}_*}^{\mathcal{Z}^{[i]}(t)})}{\sigma_f^2+(\sigma_e^{[i]})^2}\eta(\boldsymbol{z}_*^{[i]}(t)),\left(\frac{\kappa(D_{\boldsymbol{z}_*}^{\mathcal{Z}^{[i]}(t)})}{\sigma_f^2+(\sigma_e^{[i]})^2}\right)^2(\sigma_e^{[i]})^2\right)$.
We denote the probability density function of $\check{{\mu}}_{\boldsymbol{z}_{*}|\mathcal{D}^{[i]}(t)}$ by $g(\mu)\triangleq\frac{1}{\sqrt{2\pi Var\left[\check{{\mu}}_{\boldsymbol{z}_*|\mathcal{D}^{[i]}(t)}\right]}}\exp\left(-\frac{\left(\mu-\mathbb{E}\left[\check{{\mu}}_{\boldsymbol{z}_*|\mathcal{D}^{[i]}(t)}\right]\right)^2}{2Var\left[\check{{\mu}}_{\boldsymbol{z}_*|\mathcal{D}^{[i]}(t)}\right]}\right)$. The expectation of $\exp({\lambda_i(\check{{\mu}}_{\boldsymbol{z}_{*}|\mathcal{D}^{[i]}(t)}-\mathbb{E}\left[\check{{\mu}}_{\boldsymbol{z}_*|\mathcal{D}^{[i]}(t)}\right])})$ is given as	
\begin{align*}
&\mathbb{E}\left[\exp({\lambda_i(\check{{\mu}}_{\boldsymbol{z}_{*}|\mathcal{D}^{[i]}(t)}-\mathbb{E}\left[\check{{\mu}}_{\boldsymbol{z}_*|\mathcal{D}^{[i]}(t)}\right])})\right]\nonumber \\
&=\int_{-\infty}^{+\infty}\exp\left(\lambda_i\left(\mu-\mathbb{E}\left[\check{{\mu}}_{\boldsymbol{z}_*|\mathcal{D}^{[i]}(t)}\right]\right)\right)g(\mu)d\mu.
\end{align*}
Manipulating algebraic calculations renders 
\begin{align*}
&\mathbb{E}\left[\exp\left(\lambda_i(\check{{\mu}}_{\boldsymbol{z}_{*}|\mathcal{D}^{[i]}(t)}-\mathbb{E}\left[\check{{\mu}}_{\boldsymbol{z}_*|\mathcal{D}^{[i]}(t)}\right])\right)\right]\nonumber \\
&=\exp\left(\frac{\lambda_i^2Var\left[\check{{\mu}}_{\boldsymbol{z}_*|\mathcal{D}^{[i]}(t)}\right]}{2}\right)\int_{-\infty}^{+\infty} g'(\mu)d\mu
\end{align*}
where $g'(\mu)$ is a Gaussian probability density function with mean $\lambda_i Var\left[\check{{\mu}}_{\boldsymbol{z}_*|\mathcal{D}^{[i]}(t)}\right]+\mathbb{E}\left[\check{{\mu}}_{\boldsymbol{z}_*|\mathcal{D}^{[i]}(t)}\right]$ and variance $Var\left[\check{{\mu}}_{\boldsymbol{z}_*|\mathcal{D}^{[i]}(t)}\right]$. Due to the fact that $\int_{-\infty}^{+\infty} g'(\mu)d\mu=1$, then we have 
\begin{align}
	&\mathbb{E}\left[\exp\left(\lambda_i(\check{{\mu}}_{\boldsymbol{z}_{*}|\mathcal{D}^{[i]}(t)}-\mathbb{E}\left[\check{{\mu}}_{\boldsymbol{z}_*|\mathcal{D}^{[i]}(t)}\right])\right)\right]\nonumber \\
	&=\exp\left(\frac{\lambda_i^2Var\left[\check{{\mu}}_{\boldsymbol{z}_*|\mathcal{D}^{[i]}(t)}\right]}{2}\right).\label{eq:10}
\end{align}
Recall the definition that $\sigma^2=\left(\frac{\sigma_f^2\sigma_e^{\max}}{\sigma_f^2+(\sigma_e^{\min})^2}\right)^2$. By Assumption \ref{Asu:2}, it holds that the kernel function $\kappa(\cdot)$ is monotonically decreasing and $\kappa(0)=\sigma_f^2$. Therefore, we have  $Var\left[\check{{\mu}}_{\boldsymbol{z}_*|\mathcal{D}^{[i]}(t)}\right]=\left(\frac{\kappa(D_{\boldsymbol{z}_*}^{\mathcal{Z}^{[i]}(t)})}{\sigma_f^2+(\sigma_e^{[i]})^2}\sigma_e^{[i]}\right)^2\le\left(\frac{\sigma_f^2\sigma_e^{\max}}{\sigma_f^2+(\sigma_e^{\min})^2}\right)^2=\sigma^2$. Substituting $Var\left[\check{{\mu}}_{\boldsymbol{z}_*|\mathcal{D}^{[i]}(t)}\right]\le\sigma^2$ into (\ref{eq:10}) yields $\mathbb{E}\left[\exp\left(\lambda_i(\check{{\mu}}_{\boldsymbol{z}_{*}|\mathcal{D}^{[i]}(t)}-\mathbb{E}\left[\check{{\mu}}_{\boldsymbol{z}_*|\mathcal{D}^{[i]}(t)}\right])\right)\right]\le \exp\left(\frac{\lambda_i^2\sigma^2}{2}\right)$. Thus by Definition 1, we conclude that  $\check{{\mu}}_{\boldsymbol{z}_*|\mathcal{D}^{[i]}(t)}$ is a sub-Gaussian random variable with parameter $\sigma$.
$\hfill\blacksquare$

\begin{Lem}\label{Lem:5}
	Suppose Assumptions \ref{Asu:2} and \ref{Asu:3} hold. For $\boldsymbol{z}_*\in{\mathcal{Z}_*}$, with probability at least $1-\delta$, it holds that $\left|\check{{\mu}}_{\boldsymbol{z}_{*}|\mathcal{D}^{[i]}(t)}-\eta(\boldsymbol{z}_*)\right|\le\Theta(d^{\max}(t))$.
\end{Lem}
\emph{Proof:} Pick any $i\in\mathcal{V}$.
Recall that $\check{{\mu}}_{\boldsymbol{z}_{*}|\mathcal{D}^{[i]}(t)}=\frac{\kappa(D_{\boldsymbol{z}_*}^{\mathcal{Z}^{[i]}(t)})}{\sigma_f^2+(\sigma_e^{[i]})^2}{y}_{\boldsymbol{z}_*^{[i]}(t)}^{[i]}$. We have $\check{{\mu}}_{\boldsymbol{z}_{*}|\mathcal{D}^{[i]}(t)}-\eta(\boldsymbol{z}_*)=(1-\frac{\kappa(D_{\boldsymbol{z}_*}^{\mathcal{Z}^{[i]}(t)})}{\sigma_f^2+(\sigma_e^{[i]})^2})(-\eta(\boldsymbol{z}_*))+\frac{\kappa(D_{\boldsymbol{z}_*}^{\mathcal{Z}^{[i]}(t)})}{\sigma_f^2+(\sigma_e^{[i]})^2}({y}_{\boldsymbol{z}_*^{[i]}(t)}^{[i]}-\eta(\boldsymbol{z}_*^{[i]}(t)))+\frac{\kappa(D_{\boldsymbol{z}_*}^{\mathcal{Z}^{[i]}(t)})}{\sigma_f^2+(\sigma_e^{[i]})^2}(\eta(\boldsymbol{z}_*^{[i]}(t))-\eta(\boldsymbol{z}_*))$, which implies
\begin{align}
&|\check{{\mu}}_{\boldsymbol{z}_{*}|\mathcal{D}^{[i]}(t)}-\eta(\boldsymbol{z}_*)|\le\frac{\kappa(D_{\boldsymbol{z}_*}^{\mathcal{Z}^{[i]}(t)})}{\sigma_f^2+(\sigma_e^{[i]})^2}|\eta(\boldsymbol{z}_*^{[i]}(t))-\eta(\boldsymbol{z}_*)|\nonumber \\
&+\frac{\kappa(D_{\boldsymbol{z}_*}^{\mathcal{Z}^{[i]}(t)})}{\sigma_f^2+(\sigma_e^{[i]})^2}|{y}_{\boldsymbol{z}_*^{[i]}(t)}^{[i]}-\eta(\boldsymbol{z}_*^{[i]}(t))|\nonumber \\
&+(1-\frac{\kappa(D_{\boldsymbol{z}_*}^{\mathcal{Z}^{[i]}(t)})}{\sigma_f^2+(\sigma_e^{[i]})^2})|\eta(\boldsymbol{z}_*)|.\label{eq:15}
\end{align}
We derive the upper bound for each term on the right-hand side of the inequality (\ref{eq:15}). 

\emph{Term 1}. Recall that $\boldsymbol{z}_*^{[i]}(t)\in~\text{proj}(\boldsymbol{z}_*,\mathcal{Z}^{[i]}(t))$. The Lipschitz continuity of $\eta$ in Assumption \ref{Asu:3} gives
\begin{align}
&\frac{\kappa(D_{\boldsymbol{z}_*}^{\mathcal{Z}^{[i]}(t)})}{\sigma_f^2+(\sigma_e^{[i]})^2}\left|\eta(\boldsymbol{z}_*^{[i]}(t))-\eta(\boldsymbol{z}_*)\right|\le\frac{\kappa(D_{\boldsymbol{z}_*}^{\mathcal{Z}^{[i]}(t)})}{\sigma_f^2+(\sigma_e^{[i]})^2}\ell_{\eta}D_{\boldsymbol{z}_*}^{\mathcal{Z}^{[i]}(t)}\nonumber \\
&\le\frac{\sigma_f^2}{\sigma_f^2+(\sigma_e^{\min})^2}\ell_{\eta} d^{\max}(t).\label{eq:16}
\end{align}
\emph{Term 2}. By Lemma 3, we have that $\check{{\mu}}_{\boldsymbol{z}_*|\mathcal{D}^{[i]}(t)}$ is a sub-Gaussian random variable. Then by concentration inequality of the sub-Gaussian random variable (see Lemma 1.3 of \cite{RPHJ2015}), for any $\epsilon_2>0$, we have
\begin{align*}
P\left\{\left|\check{{\mu}}_{\boldsymbol{z}_*|\mathcal{D}^{[i]}(t)}-\frac{\kappa(D_{\boldsymbol{z}_*}^{\mathcal{Z}^{[i]}(t)})}{\sigma_f^2+(\sigma_e^{[i]})^2}\eta(\boldsymbol{z}_*^{[i]}(t))\right|>\epsilon_2 \right\}\le 2e^{-\frac{\epsilon_2^2}{2\sigma^2}}.
\end{align*}
Combining the above inequality with $\check{{\mu}}_{\boldsymbol{z}_{*}|\mathcal{D}^{[i]}(t)}=\frac{\kappa(D_{\boldsymbol{z}_*}^{\mathcal{Z}^{[i]}(t)})}{\sigma_f^2+(\sigma_e^{[i]})^2}{y}_{\boldsymbol{z}_*^{[i]}(t)}^{[i]}$, for $0<\delta<1$, choosing $\epsilon_2\triangleq\sqrt{2\sigma^2(\ln2-\ln\delta)}$, with probability at least $1-\delta$, it holds
\begin{align}
&\frac{\kappa(D_{\boldsymbol{z}_*}^{\mathcal{Z}^{[i]}(t)})}{\sigma_f^2+(\sigma_e^{[i]})^2}|{y}_{\boldsymbol{z}_*^{[i]}(t)}^{[i]}-\eta(\boldsymbol{z}_*^{[i]}(t))|\nonumber \\
&=\left|\check{{\mu}}_{\boldsymbol{z}_*|\mathcal{D}^{[i]}(t)}-\frac{\kappa(D_{\boldsymbol{z}_*}^{\mathcal{Z}^{[i]}(t)})}{\sigma_f^2+(\sigma_e^{[i]})^2}\eta(\boldsymbol{z}_*^{[i]}(t))\right|\nonumber \\
&\le\sqrt{2\sigma^2(\ln2-\ln\delta)}.\label{eq:17}
\end{align}
\emph{Term 3}. By monotonicity of $\kappa(\cdot)$ in Assumption \ref{Asu:2}, it gives
\begin{align}\label{eq:18}
(1-\frac{\kappa(D_{\boldsymbol{z}_*}^{\mathcal{Z}^{[i]}(t)})}{\sigma_f^2+(\sigma_e^{[i]})^2})|\eta(\boldsymbol{z}_*)|\le(1-\frac{\kappa(d^{\max}(t))}{\sigma_f^2+(\sigma_e^{\max})^2})\|\eta\|_{\infty}.
\end{align}
Therefore, applying the inequalities (\ref{eq:16}), (\ref{eq:17}) and (\ref{eq:18}) to (\ref{eq:15}), for $0<\delta<1$, with probability at least $1-\delta$, we have that for all $i\in\mathcal{I}(t)$, the desired inequality holds.
$\hfill\blacksquare$

{{
\begin{Lem}\label{Lem:55}
	Suppose Assumptions \ref{Asu:00}, \ref{Asu:2} and \ref{Asu:3} hold. For $\boldsymbol{z}_*\in{\mathcal{Z}_*}$ and $i\in\mathcal{I}(t)$, with probability at least $1-\delta$, it holds that $\left|\check{{\mu}}'_{\boldsymbol{z}_{*}|\mathcal{D}^{[i]}(t)}-\eta(\boldsymbol{z}_*)\right|\le\Theta(d^{\max}(t))$.
\end{Lem}
\emph{Proof:} Pick any $i \in \mathcal{I}(t)$, it follows from Lemma \ref{Lem:0} that 
\begin{align}\label{eq:200}
&\left|\check{{\mu}}'_{\boldsymbol{z}_{*}|\mathcal{D}^{[i]}(t)}-\eta(\boldsymbol{z}_*)\right|\le\max\left\{\left|\min_{j\in\mathcal{V}\backslash\mathcal{B}}\left\{\check{{\mu}}_{\boldsymbol{z}_{*}|\mathcal{D}^{[j]}(t)}\right\}-\eta(\boldsymbol{z}_*)\right|,\right. \nonumber \\
&\left.\left|\max_{j\in\mathcal{V}\backslash\mathcal{B}}\left\{\check{{\mu}}_{\boldsymbol{z}_{*}|\mathcal{D}^{[j]}(t)}\right\}-\eta(\boldsymbol{z}_*)\right|\right\}.
\end{align}
By Lemma \ref{Lem:5}, with probability at least $1-\delta$, we have  $\left|\check{{\mu}}_{\boldsymbol{z}_{*}|\mathcal{D}^{[i]}(t)}-\eta(\boldsymbol{z}_*)\right|\le\Theta(d^{\max}(t))$ for all $i\in\mathcal{V}$. Then for $i\in\mathcal{I}(t)$, the inequality (\ref{eq:200}) becomes $\left|\check{{\mu}}'_{\boldsymbol{z}_{*}|\mathcal{D}^{[i]}(t)}-\eta(\boldsymbol{z}_*)\right|\le\Theta(d^{\max}(t))$.
}}
$\hfill\blacksquare$

\begin{Lem}\label{Lem:4}
	Let Assumptions \ref{Asu:00} and \ref{Asu:2} hold. For all $\boldsymbol{z}_*\in{\mathcal{Z}_*}$ and  $0<\delta<1$, with probability at least $1-\delta$, it holds that
	 $\frac{\hat{\sigma}^2_{\boldsymbol{z}_*|\mathcal{D}(t)}}{|\mathcal{I}(t)|}\sum_{i\in\mathcal{I}(t)}\check{\sigma}'^{-2}_{\boldsymbol{z}_{*}|\mathcal{D}^{[i]}(t)}\left|\check{{\mu}}'_{\boldsymbol{z}_{*}|\mathcal{D}^{[i]}(t)}-\check{{\mu}}_{\boldsymbol{z}_{*}|\mathcal{D}^{[i]}(t)}\right|\le\Delta(d^{\max}(t))$.
\end{Lem}
\emph{Proof:} We denote by $\mathcal{F}(t)\triangleq\mathcal{I}(t)\bigcap(\mathcal{V}\backslash\mathcal{B})$ the set of benign agents in the set $\mathcal{I}(t)$. Notice that, $\check{{\mu}}'_{\boldsymbol{z}_{*}|\mathcal{D}^{[i]}(t)}=\check{{\mu}}_{\boldsymbol{z}_{*}|\mathcal{D}^{[i]}(t)}$ for all $i\in\mathcal{F}(t)$. We have
\begin{align}
&\frac{\hat{\sigma}^2_{\boldsymbol{z}_*|\mathcal{D}(t)}}{|\mathcal{I}(t)|}\sum_{i\in\mathcal{I}(t)}\check{\sigma}'^{-2}_{\boldsymbol{z}_{*}|\mathcal{D}^{[i]}(t)}\left|\check{{\mu}}'_{\boldsymbol{z}_{*}|\mathcal{D}^{[i]}(t)}-\check{{\mu}}_{\boldsymbol{z}_{*}|\mathcal{D}^{[i]}(t)}\right|\nonumber \\
&=\frac{\hat{\sigma}^2_{\boldsymbol{z}_*|\mathcal{D}(t)}}{|\mathcal{I}(t)|}\sum_{i\in\mathcal{I}(t)\bigcap\mathcal{B}}\check{\sigma}'^{-2}_{\boldsymbol{z}_{*}|\mathcal{D}^{[i]}(t)}\left|\check{{\mu}}'_{\boldsymbol{z}_{*}|\mathcal{D}^{[i]}(t)}-\check{{\mu}}_{\boldsymbol{z}_{*}|\mathcal{D}^{[i]}(t)}\right|\nonumber \\
&\quad+\underbrace{\frac{\hat{\sigma}^2_{\boldsymbol{z}_*|\mathcal{D}(t)}}{|\mathcal{I}(t)|}\sum_{i\in\mathcal{F}(t)}\check{\sigma}'^{-2}_{\boldsymbol{z}_{*}|\mathcal{D}^{[i]}(t)}\left|\check{{\mu}}'_{\boldsymbol{z}_{*}|\mathcal{D}^{[i]}(t)}-\check{{\mu}}_{\boldsymbol{z}_{*}|\mathcal{D}^{[i]}(t)}\right|}_{=0}\nonumber \\
& = \frac{\sum_{i\in\mathcal{I}(t)\bigcap\mathcal{B}}\check{\sigma}'^{-2}_{\boldsymbol{z}_{*}|\mathcal{D}^{[i]}(t)}\left|\check{{\mu}}'_{\boldsymbol{z}_{*}|\mathcal{D}^{[i]}(t)}-\check{{\mu}}_{\boldsymbol{z}_{*}|\mathcal{D}^{[i]}(t)}\right|}{\sum_{j\in\mathcal{I}(t)}\check{\sigma}'^{-2}_{\boldsymbol{z}_{*}|\mathcal{D}^{[j]}(t)}}\label{eq:11} 
\end{align}
where  $\hat{\sigma}^2_{\boldsymbol{z}_*|\mathcal{D}(t)}=\frac{|\mathcal{I}(t)|}{\sum_{j\in\mathcal{I}(t)}\check{\sigma}'^{-2}_{\boldsymbol{z}_{*}|\mathcal{D}^{[j]}(t)}}$ is used.
{{By triangle inequality, Lemma \ref{Lem:5} and Lemma \ref{Lem:55}, we have  $\left|\check{{\mu}}'_{\boldsymbol{z}_{*}|\mathcal{D}^{[i]}(t)}-\check{{\mu}}_{\boldsymbol{z}_{*}|\mathcal{D}^{[i]}(t)}\right|\le2\Theta(d^{\max}(t))$.}}  

In the remaining proof, we characterize the lower and upper bounds of $\check{\sigma}'^{-2}_{\boldsymbol{z}_{*}|\mathcal{D}^{[i]}(t)}$.
Pick any $i\in\mathcal{I}(t)$, Lemma 2 renders that
\begin{align*}
	\min_{j\in\mathcal{V}\backslash\mathcal{B}}\left\{\check{\sigma}^{2}_{\boldsymbol{z}_{*}|\mathcal{D}^{[j]}(t)}\right\}\le\check{\sigma}'^{2}_{\boldsymbol{z}_{*}|\mathcal{D}^{[i]}(t)}\le\max_{j\in\mathcal{V}\backslash\mathcal{B}}\left\{\check{\sigma}^{2}_{\boldsymbol{z}_{*}|\mathcal{D}^{[j]}(t)}\right\},
\end{align*}
then we have
\begin{align*}
	\left(\max_{j\in\mathcal{V}\backslash\mathcal{B}}\left\{\check{\sigma}^{2}_{\boldsymbol{z}_{*}|\mathcal{D}^{[j]}(t)}\right\}\right)^{-1}&\le\check{\sigma}'^{-2}_{\boldsymbol{z}_{*}|\mathcal{D}^{[i]}(t)} \\
	&\le\left(\min_{j\in\mathcal{V}\backslash\mathcal{B}}\left\{\check{\sigma}^{2}_{\boldsymbol{z}_{*}|\mathcal{D}^{[j]}(t)}\right\}\right)^{-1}.
\end{align*}
Theorem IV.3 in \cite{ZM2021} gives $\frac{\sigma_f^2(\sigma_e^{[i]})^2}{\sigma_f^2+(\sigma_e^{[i]})^2}\le\check{\sigma}_{\boldsymbol{z}_*|\mathcal{D}^{[i]}(t)}^2 \le\sigma_f^2-\frac{\kappa(d^{[i]}(t))^2}{\sigma_f^2+(\sigma_e^{[i]})^2}$ for all $\boldsymbol{z}_*\in\mathcal{Z}_*$. By monotonicity of $\kappa(\cdot)$ in Assumption \ref{Asu:2}, it holds that $\frac{\sigma_f^2(\sigma_e^{\min})^2}{\sigma_f^2+(\sigma_e^{\max})^2}\le\check{\sigma}_{\boldsymbol{z}_*|\mathcal{D}^{[i]}(t)}^2\le\sigma_f^2-\frac{\kappa(d^{\max}(t))^2}{\sigma_f^2+(\sigma_e^{\max})^2}$, which implies that for any $\boldsymbol{z}_{*}\in{\mathcal{Z}_*}$, we have $\frac{\sigma_f^2+(\sigma_e^{\max})^2}{\sigma_f^4+\sigma_f^2(\sigma_e^{\max})^2-\kappa(d^{\max}(t))^2}\le\check{\sigma}'^{-2}_{\boldsymbol{z}_{*}|\mathcal{D}^{[i]}(t)}\le\frac{\sigma_f^2+(\sigma_e^{\max})^2}{\sigma_f^2(\sigma_e^{\min})^2}$.
{{First,}} Lemma 1 shows that $|\mathcal{I}(t)|\ge n-4\beta n$. Then we have 
\begin{eqnarray}\label{eq:13}
\sum_{j\in\mathcal{I}(t)}\check{\sigma}'^{-2}_{\boldsymbol{z}_{*}|\mathcal{D}^{[j]}(t)}\ge\frac{(1-4\beta)n\left(\sigma_f^2+(\sigma_e^{\max})^2\right)}{\sigma_f^4+\sigma_f^2(\sigma_e^{\max})^2-\kappa(d^{\max}(t))^2}.
\end{eqnarray}

{{Second,}} since $|\mathcal{I}(t)\bigcap\mathcal{B}|\le|\mathcal{B}|=\alpha n$, then with probability at least $1-\delta$, we have
\begin{align}
&\sum_{i\in\mathcal{I}(t)\bigcap\mathcal{B}}\check{\sigma}'^{-2}_{\boldsymbol{z}_{*}|\mathcal{D}^{[i]}(t)}\left(\left|\check{{\mu}}'_{\boldsymbol{z}_{*}|\mathcal{D}^{[i]}(t)}-\check{{\mu}}_{\boldsymbol{z}_{*}|\mathcal{D}^{[i]}(t)}\right|\right)\nonumber \\
&\le2\alpha n\frac{\sigma_f^2+(\sigma_e^{\max})^2}{\sigma_f^2(\sigma_e^{\min})^2}{{\Theta(d^{\max}(t))}}.\label{eq:14}
\end{align}
{{Therefore, combining}} (\ref{eq:13}) and (\ref{eq:14}) with (\ref{eq:11}) renders that with probability at least $1-\delta$, the desired inequality holds.
$\hfill\blacksquare$

With Lemmas \ref{Lem:3}, \ref{Lem:5}, {{\ref{Lem:55}}} and \ref{Lem:4}, we now proceed to complete the proof of part I in Theorem \ref{Thm:2}.
Notice that
$\frac{\hat{\sigma}^2_{\boldsymbol{z}_*|\mathcal{D}(t)}}{|\mathcal{I}(t)|}\sum_{i\in\mathcal{I}(t)}\check{\sigma}'^{-2}_{\boldsymbol{z}_{*}|\mathcal{D}^{[i]}(t)}=1$, by {{triangle}} inequality, we have
\begin{align*}
&\left|\frac{\hat{\sigma}^2_{\boldsymbol{z}_*|\mathcal{D}(t)}}{|\mathcal{I}(t)|}\sum_{i\in\mathcal{I}(t)}\check{{\mu}}'_{\boldsymbol{z}_{*}|\mathcal{D}^{[i]}(t)}\check{\sigma}'^{-2}_{\boldsymbol{z}_{*}|\mathcal{D}^{[i]}(t)}-\eta(\boldsymbol{z}_*)\right| \nonumber \\
&=\left|\frac{\hat{\sigma}^2_{\boldsymbol{z}_*|\mathcal{D}(t)}}{|\mathcal{I}(t)|}\sum_{i\in\mathcal{I}(t)}\check{\sigma}'^{-2}_{\boldsymbol{z}_{*}|\mathcal{D}^{[i]}(t)}\left(\check{{\mu}}'_{\boldsymbol{z}_{*}|\mathcal{D}^{[i]}(t)}-\eta(\boldsymbol{z}_*)\right)\right| \nonumber \\
&\le\frac{\hat{\sigma}^2_{\boldsymbol{z}_*|\mathcal{D}(t)}}{|\mathcal{I}(t)|}\sum_{i\in\mathcal{I}(t)}\check{\sigma}'^{-2}_{\boldsymbol{z}_{*}|\mathcal{D}^{[i]}(t)}\left|\check{{\mu}}_{\boldsymbol{z}_{*}|\mathcal{D}^{[i]}(t)}-\eta(\boldsymbol{z}_*)\right| \nonumber \\
&\quad+\frac{\hat{\sigma}^2_{\boldsymbol{z}_*|\mathcal{D}(t)}}{|\mathcal{I}(t)|}\sum_{i\in\mathcal{I}(t)}\check{\sigma}'^{-2}_{\boldsymbol{z}_{*}|\mathcal{D}^{[i]}(t)}\left|\check{{\mu}}'_{\boldsymbol{z}_{*}|\mathcal{D}^{[i]}(t)}-\check{{\mu}}_{\boldsymbol{z}_{*}|\mathcal{D}^{[i]}(t)}\right|. 
\end{align*}
By (\ref{eq:7}), we have $0<\frac{\hat{\sigma}^2_{\boldsymbol{z}_*|\mathcal{D}(t)}}{|\mathcal{I}(t)|}\check{\sigma}'^{-2}_{\boldsymbol{z}_{*}|\mathcal{D}^{[i]}(t)}<1$ and  $\frac{\hat{\sigma}^2_{\boldsymbol{z}_*|\mathcal{D}(t)}}{|\mathcal{I}(t)|}\sum_{i\in\mathcal{I}(t)}\check{\sigma}'^{-2}_{\boldsymbol{z}_{*}|\mathcal{D}^{[i]}(t)}=1$. By Lemma \ref{Lem:5}, with probability at least $1-\delta$, it holds that $\frac{\hat{\sigma}^2_{\boldsymbol{z}_*|\mathcal{D}(t)}}{|\mathcal{I}(t)|}\sum_{i\in\mathcal{I}(t)}\check{\sigma}'^{-2}_{\boldsymbol{z}_{*}|\mathcal{D}^{[i]}(t)}\notag\\\left|\check{{\mu}}_{\boldsymbol{z}_{*}|\mathcal{D}^{[i]}(t)}-\eta(\boldsymbol{z}_*)\right|\le\Theta(d^{\max}(t))$. Therefore, we complete the proof of part I.

\textbf{Part II:} We give the upper bound and lower bound of $\hat{\sigma}'^2_{\boldsymbol{z}_*|\mathcal{D}(t)}$ as follows:

1) Upper bound. Recall that $\hat{\sigma}^2_{\boldsymbol{z}_*|\mathcal{D}(t)}=\frac{|\mathcal{I}(t)|}{\sum_{i\in\mathcal{I}(t)}\check{\sigma}'^{-2}_{\boldsymbol{z}_{*}|\mathcal{D}^{[i]}(t)}}$. Note that $f(x)=\frac{1}{x}$ is a convex function for $x>0$. By Jensen's inequality (see page 21 in \cite{TT2012}), we have $f(\frac{1}{n}\sum_{i=1}^nx_i)\le\frac{1}{n}\sum_{i=1}^nf(x_i)$. Then plugging in $x_i=\check{\sigma}'^{-2}_{\boldsymbol{z}_{*}|\mathcal{D}^{[i]}(t)}$, we have
\begin{align}\label{eq:19}
\hat{\sigma}^2_{\boldsymbol{z}_*|\mathcal{D}(t)}&=\frac{|\mathcal{I}(t)|}{\sum_{i\in\mathcal{I}(t)}\check{\sigma}'^{-2}_{\boldsymbol{z}_{*}|\mathcal{D}^{[i]}(t)}}=f(\frac{\sum_{i\in\mathcal{I}(t)}\check{\sigma}'^{-2}_{\boldsymbol{z}_{*}|\mathcal{D}^{[i]}(t)}}{|\mathcal{I}(t)|})\nonumber \\
&\le\frac{1}{|\mathcal{I}(t)|}\sum_{i\in\mathcal{I}(t)}\check{\sigma}'^{2}_{\boldsymbol{z}_{*}|\mathcal{D}^{[i]}(t)}.
\end{align}
It suffices to show the upper bound of $\sum_{i\in\mathcal{I}(t)}\check{\sigma}'^{2}_{\boldsymbol{z}_{*}|\mathcal{D}^{[i]}(t)}$. Recall that we decompose the agent set $\mathcal{I}(t)$ into two subsets $\mathcal{F}(t)$ and $\mathcal{I}(t)\bigcap\mathcal{B}$ where $\mathcal{F}(t)=\mathcal{I}(t)\bigcap(\mathcal{V}\backslash\mathcal{B})$ contains the benign agents and $\mathcal{I}(t)\bigcap\mathcal{B}$ contains the Byzantine agents, then we have
\begin{align} \label{eq:20}
\sum_{i\in\mathcal{I}(t)}\check{\sigma}'^{2}_{\boldsymbol{z}_{*}|\mathcal{D}^{[i]}(t)}=\sum_{i\in\mathcal{F}(t)}\check{\sigma}'^{2}_{\boldsymbol{z}_{*}|\mathcal{D}^{[i]}(t)}+\sum_{i\in\mathcal{I}(t)\bigcap\mathcal{B}}\check{\sigma}'^{2}_{\boldsymbol{z}_{*}|\mathcal{D}^{[i]}(t)}.
\end{align}
We proceed to analyze each term of the above equation.

First, notice that $\mathcal{F}(t)$ is the set of benign agents, hence it holds that $\check{\sigma}'^{2}_{\boldsymbol{z}_{*}|\mathcal{D}^{[i]}(t)}=\check{\sigma}^{2}_{\boldsymbol{z}_{*}|\mathcal{D}^{[i]}(t)}$ for all $i\in\mathcal{F}(t)$. According to Theorem IV.3 in \cite{ZM2021}, we have $\check{\sigma}_{\boldsymbol{z}_*|\mathcal{D}^{[i]}(t)}^2 \le \sigma_f^2-\frac{\kappa(d^{[i]}(t))^2}{\sigma_f^2+(\sigma_e^{[i]})^2}$ for all $\boldsymbol{z}_*\in\mathcal{Z}_*$. Monotonicity of $\kappa(\cdot)$ in Assumption \ref{Asu:2} shows that $\check{\sigma}^{2}_{\boldsymbol{z}_{*}|\mathcal{D}^{[i]}(t)}\le\sigma_f^2-\frac{\kappa(d^{\max}(t))^2}{\sigma_f^2+(\sigma_e^{\max})^2}$, {{which implies
\begin{align}\label{eq:21}
	\sum_{i\in\mathcal{F}(t)}\check{\sigma}'^{2}_{\boldsymbol{z}_{*}|\mathcal{D}^{[i]}(t)}\le(1-\alpha)n\left(\sigma_f^2-\frac{\kappa(d^{\max}(t))^2}{\sigma_f^2+(\sigma_e^{\max})^2}\right).
\end{align}		
}}


{{Second,}} for $i\in\mathcal{I}(t)\bigcap\mathcal{B}$, it follows from Lemma \ref{Lem:0} that $\check{\sigma}'^{2}_{\boldsymbol{z}_{*}|\mathcal{D}^{[i]}(t)}\le\max_{i\in\mathcal{V}\backslash\mathcal{B}}\left\{\check{\sigma}^{2}_{\boldsymbol{z}_{*}|\mathcal{D}^{[i]}(t)}\right\}\le\sigma_f^2-\frac{\kappa(d^{\max}(t))^2}{\sigma_f^2+(\sigma_e^{\max})^2}$, {{which implies
\begin{align}\label{eq:22}
	\sum_{i\in\mathcal{I}(t)\bigcap\mathcal{B}}\check{\sigma}'^{2}_{\boldsymbol{z}_{*}|\mathcal{D}^{[i]}(t)}\le\alpha n\left(\sigma_f^2-\frac{\kappa(d^{\max}(t))^2}{\sigma_f^2+(\sigma_e^{\max})^2}\right).
\end{align}		
}}

Therefore, {{combining (\ref{eq:21}) and (\ref{eq:22}) with (\ref{eq:20}) gives $\sum_{i\in\mathcal{I}(t)}\check{\sigma}'^{2}_{\boldsymbol{z}_{*}|\mathcal{D}^{[i]}(t)}\le n\left(\sigma_f^2-\frac{\kappa(d^{\max}(t))^2}{\sigma_f^2+(\sigma_e^{\max})^2}\right)$. By the inequality (\ref{eq:19}) and Lemma \ref{Lem:22}}}, the upper bound of $\hat{\sigma}^2_{\boldsymbol{z}_*|\mathcal{D}(t)}$ is given as 
\begin{align}\label{eq:23}
\hat{\sigma}^2_{\boldsymbol{z}_*|\mathcal{D}(t)}\le{{\frac{1}{1-4\beta}}}\left(\sigma_f^2-\frac{\kappa(d^{\max}(t))^2}{\sigma_f^2+(\sigma_e^{\max})^2}\right).
\end{align} 
2) Lower bound. Similar to the analysis of upper bound, we have
\begin{align}\label{eq:24}
\sum_{i\in\mathcal{I}(t)}\check{\sigma}'^{-2}_{\boldsymbol{z}_{*}|\mathcal{D}^{[i]}(t)}=\sum_{i\in\mathcal{F}(t)}\check{\sigma}'^{-2}_{\boldsymbol{z}_{*}|\mathcal{D}^{[i]}(t)}+\sum_{i\in\mathcal{I}(t)\bigcap\mathcal{B}}\check{\sigma}'^{-2}_{\boldsymbol{z}_{*}|\mathcal{D}^{[i]}(t)}.
\end{align}
{{First}}, for all $i\in\mathcal{F}(t)$, it holds that $\check{\sigma}'^{-2}_{\boldsymbol{z}_{*}|\mathcal{D}^{[i]}(t)}=\check{\sigma}^{-2}_{\boldsymbol{z}_{*}|\mathcal{D}^{[i]}(t)}$. Theorem IV.3 in \cite{ZM2021} gives $\check{\sigma}_{\boldsymbol{z}_*|\mathcal{D}^{[i]}(t)}^2 \ge\frac{\sigma_f^2(\sigma_e^{[i]})^2}{\sigma_f^2+(\sigma_e^{[i]})^2}$ for all $\boldsymbol{z}_*\in\mathcal{Z}_*$, which implies $\check{\sigma}_{\boldsymbol{z}_*|\mathcal{D}^{[i]}(t)}^{-2}\le\frac{\sigma_f^2+(\sigma_e^{\max})^2}{\sigma_f^2(\sigma_e^{\min})^2}$. {{We have
\begin{align}\label{eq:25}
	\sum_{i\in\mathcal{F}(t)}\check{\sigma}_{\boldsymbol{z}_*|\mathcal{D}^{[i]}(t)}'^{-2}\le(1-\alpha)n\frac{\sigma_f^2+(\sigma_e^{\max})^2}{\sigma_f^2(\sigma_e^{\min})^2}.
\end{align}		
}}

{{Second}}, for $i\in\mathcal{I}(t)\bigcap\mathcal{B}$, Lemma 2 implies the following inequality $\check{\sigma}'^{-2}_{\boldsymbol{z}_{*}|\mathcal{D}^{[i]}(t)}\le \left(\min_{i\in\mathcal{V}\backslash\mathcal{B}}\left\{\check{\sigma}^{2}_{\boldsymbol{z}_{*}|\mathcal{D}^{[i]}(t)}\right\}\right)^{-1} \le\frac{\sigma_f^2+(\sigma_e^{\max})^2}{\sigma_f^2(\sigma_e^{\min})^2}$. {{We have
\begin{align}\label{eq:26}
	\sum_{i\in\mathcal{I}(t)\bigcap\mathcal{B}}\check{\sigma}_{\boldsymbol{z}_*|\mathcal{D}^{[i]}(t)}'^{-2}\le\alpha n\frac{\sigma_f^2+(\sigma_e^{\max})^2}{\sigma_f^2(\sigma_e^{\min})^2}
\end{align}		
}}

Therefore, {{combining (\ref{eq:25}) and (\ref{eq:26}) with (\ref{eq:24}) gives}}  $\sum_{i\in\mathcal{I}(t)}\check{\sigma}'^{-2}_{\boldsymbol{z}_{*}|\mathcal{D}^{[i]}(t)}\le n\frac{\sigma_f^2+(\sigma_e^{\max})^2}{\sigma_f^2(\sigma_e^{\min})^2}$. Recall that $\hat{\sigma}^2_{\boldsymbol{z}_*|\mathcal{D}(t)}=\frac{|\mathcal{I}(t)|}{\sum_{i\in\mathcal{I}(t)}\check{\sigma}'^{-2}_{\boldsymbol{z}_{*}|\mathcal{D}^{[i]}(t)}}$, then by Lemma \ref{Lem:22}, we have
\begin{align}\label{eq:27}
\hat{\sigma}^2_{\boldsymbol{z}_*|\mathcal{D}(t)}\ge{{(1-2\beta)}}\left(\frac{\sigma_f^2(\sigma_e^{\min})^2}{\sigma_f^2+(\sigma_e^{\max})^2}\right).
\end{align}
Thus, combining (\ref{eq:23}) and (\ref{eq:27}), the proof is complete.

\subsection{Proof of Theorem \ref{Thm:3}}
\textbf{Part I:} Pick any $i\in\mathcal{V}\backslash\mathcal{B}$. If $\boldsymbol{z}_*\in\mathcal{Z}_{fused}^{[i]}(t)$, we have $\tilde{{\mu}}^{[i]}_{\boldsymbol{z}_*|\mathcal{D}(t)}= \hat{\mu}_{\boldsymbol{z}_{*}|\mathcal{D}(t)}$, and
$(\tilde{\sigma}^{[i]}_{\boldsymbol{z}_*|\mathcal{D}(t)})^{2}=\hat{\sigma}^2_{\boldsymbol{z}_{*}|\mathcal{D}(t)}$. By Part I of Theorem \ref{Thm:2}, with probability at least $1-\delta$, we have $|\tilde{{\mu}}^{[i]}_{\boldsymbol{z}_*|\mathcal{D}(t)}-\eta(\boldsymbol{z}_*)|\le\Theta(d^{\max}(t))+\Delta(d^{\max}(t))$. 

If $\boldsymbol{z}_*\notin\mathcal{Z}_{fused}^{[i]}(t)$, we have $\tilde{{\mu}}^{[i]}_{\boldsymbol{z}_*|\mathcal{D}(t)}= \check{\mu}_{\boldsymbol{z}_{*}|\mathcal{D}^{[i]}(t)}$, and
$(\tilde{\sigma}^{[i]}_{\boldsymbol{z}_*|\mathcal{D}(t)})^{2}=\check{\sigma}^2_{\boldsymbol{z}_{*}|\mathcal{D}^{[i]}(t)}$. By Lemma \ref{Lem:5}, with probability at least $1-\delta$, we have $\left|\check{{\mu}}_{\boldsymbol{z}_{*}|\mathcal{D}^{[i]}(t)}-\eta(\boldsymbol{z}_*) \right|\le\Theta(d^{\max}(t))$. Notice that $\Delta(d^{\max}(t))>0$. Therefore, for $i\in\mathcal{V}\backslash\mathcal{B}$, the desired inequality holds with probability at least $1-\delta$.

\textbf{Part II:}
Pick any $i\in\mathcal{V}$. Recall that $\check{{\mu}}_{\boldsymbol{z}_{*}|\mathcal{D}^{[i]}(t)}=\frac{\kappa(D_{\boldsymbol{z}_*}^{\mathcal{Z}^{[i]}(t)})}{\sigma_f^2+(\sigma_e^{[i]})^2}{y}_{\boldsymbol{z}_*^{[i]}(t)}^{[i]}$. By the observation model (\ref{eq:4}), we decompose $\check{{\mu}}_{\boldsymbol{z}_{*}|\mathcal{D}^{[i]}(t)}$ into a deterministic process plus a zero-mean Gaussian process. We denote $\check{r}_{\boldsymbol{z}_*}^{[i]}(t)\triangleq\frac{\kappa(D_{\boldsymbol{z}_*}^{\mathcal{Z}^{[i]}(t)})}{\sigma_f^2+(\sigma_e^{[i]})^2}\eta(\boldsymbol{z}_*^{[i]}(t))$ and $\check{e}_{\boldsymbol{z}_*}^{[i]}(t)\triangleq\frac{\kappa(D_{\boldsymbol{z}_*}^{\mathcal{Z}^{[i]}(t)})}{\sigma_f^2+(\sigma_e^{[i]})^2}({y}_{\boldsymbol{z}_*^{[i]}(t)}^{[i]}-\eta(\boldsymbol{z}_*^{[i]}(t)))$, then we have $\check{{\mu}}_{\boldsymbol{z}_*|\mathcal{D}^{[i]}(t)}=\check{r}_{\boldsymbol{z}_*}^{[i]}(t)+\check{e}_{\boldsymbol{z}_*}^{[i]}(t)$. 

\begin{Lem}\label{Lem:66}
	It holds that	
	\begin{itemize}
		\item [1)] $\mathbb{E}[\check{r}_{\boldsymbol{z}_*}^{[i]}(t)^2]=\check{r}_{\boldsymbol{z}_*}^{[i]}(t)^2$;
		\item [2)] $\mathbb{E}[\check{e}_{\boldsymbol{z}_*}^{[i]}(t)^2]=\left(\frac{\sigma_e^{[i]}\kappa(D_{\boldsymbol{z}_*}^{\mathcal{Z}^{[i]}(t)})}{\sigma_f^2+(\sigma_e^{[i]})^2}\right)^2$;
		\item [3)] $\mathbb{E}[\check{r}_{\boldsymbol{z}_*}^{[i]}(t)\check{e}_{\boldsymbol{z}_*}^{[i]}(t)]=0$;
		\item [4)] $\mathbb{E}[\check{r}_{\boldsymbol{z}_*}^{[i]}(t)\eta(\boldsymbol{z}_*)]=\frac{\kappa(D_{\boldsymbol{z}_*}^{\mathcal{Z}^{[i]}(t)})}{\sigma_f^2+(\sigma_e^{[i]})^2}\eta(\boldsymbol{z}_*^{[i]}(t))\eta(\boldsymbol{z}_*)$;
		\item [5)] $\mathbb{E}[\check{e}_{\boldsymbol{z}_*}^{[i]}(t)\eta(\boldsymbol{z}_*)]=0$.
	\end{itemize}
\end{Lem}

\emph{Proof:} 1) Note that $\check{r}_{\boldsymbol{z}_*}^{[i]}(t)=\frac{\kappa(D_{\boldsymbol{z}_*}^{\mathcal{Z}^{[i]}(t)})}{\sigma_f^2+(\sigma_e^{[i]})^2}\eta(\boldsymbol{z}_*^{[i]}(t))$ is a deterministic process, we have $\mathbb{E}[\check{r}_{\boldsymbol{z}_*}^{[i]}(t)^2]=\check{r}_{\boldsymbol{z}_*}^{[i]}(t)^2$.

2) Since $\check{e}_{\boldsymbol{z}_*}^{[i]}(t)$ is a zero-mean Gaussian process with $\mathbb{E}[\check{e}_{\boldsymbol{z}_*}^{[i]}(t)]=0$ and $Var(\check{e}_{\boldsymbol{z}_*}^{[i]}(t))=\left(\frac{\kappa(D_{\boldsymbol{z}_*}^{\mathcal{Z}^{[i]}(t)})}{\sigma_f^2+(\sigma_e^{[i]})^2}\right)^2(\sigma_e^{[i]})^2$, then we have $\mathbb{E}[\check{e}_{\boldsymbol{z}_*}^{[i]}(t)^2]=Var(\check{e}_{\boldsymbol{z}_*}^{[i]}(t))+\mathbb{E}[\check{e}_{\boldsymbol{z}_*}^{[i]}(t)]^2=\left(\frac{\kappa(D_{\boldsymbol{z}_*}^{\mathcal{Z}^{[i]}(t)})}{\sigma_f^2+(\sigma_e^{[i]})^2}\right)^2(\sigma_e^{[i]})^2$.

3) Recall that $\mathbb{E}[\check{e}_{\boldsymbol{z}_*}^{[i]}(t)]=0$. Then we have $\mathbb{E}[\check{r}_{\boldsymbol{z}_*}^{[i]}(t)\check{e}_{\boldsymbol{z}_*}^{[i]}(t)]\notag\\=\check{r}_{\boldsymbol{z}_*}^{[i]}(t)\mathbb{E}[\check{e}_{\boldsymbol{z}_*}^{[i]}(t)]=0$.

4) Since $\check{r}_{\boldsymbol{z}_*}^{[i]}(t)=\frac{\kappa(D_{\boldsymbol{z}_*}^{\mathcal{Z}^{[i]}(t)})}{\sigma_f^2+(\sigma_e^{[i]})^2}\eta(\boldsymbol{z}_*^{[i]}(t))$ and $\check{r}_{\boldsymbol{z}_*}^{[i]}(t)$ is deterministic, we have $\mathbb{E}[\check{r}_{\boldsymbol{z}_*}^{[i]}(t)\eta(\boldsymbol{z}_*)]=\frac{\kappa(D_{\boldsymbol{z}_*}^{\mathcal{Z}^{[i]}(t)})}{\sigma_f^2+(\sigma_e^{[i]})^2}\eta(\boldsymbol{z}_*^{[i]}(t))\eta(\boldsymbol{z}_*)$.

5) The proof is similar to that of term 3, and omitted for brevity. 
$\hfill\blacksquare$

The following lemma provides a lower bound of the mean squared error for the predictive mean from the agent-based local GPR.		
\begin{Lem}\label{Lem:6}
Let Assumption \ref{Asu:2} hold. For $i\in\mathcal{V}$ and $\boldsymbol{z}_*\in\mathcal{Z}_*$, it holds that $\mathbb{E}[(\check{{\mu}}_{\boldsymbol{z}_*|\mathcal{D}^{[i]}(t)}-\eta(\boldsymbol{z}_*))^2]\ge\left(\frac{\sigma_e^{[i]}\kappa(d^{[i]}(t))}{\sigma_f^2+(\sigma_e^{[i]})^2}\right)^2$.
\end{Lem}
		
\emph{Proof:}
	Notice that
	\begin{align*}
		&(\check{{\mu}}_{\boldsymbol{z}_*|\mathcal{D}^{[i]}(t)}-\eta(\boldsymbol{z}_*))^2=(\check{r}_{\boldsymbol{z}_*}^{[i]}(t)+\check{e}_{\boldsymbol{z}_*}^{[i]}(t)-\eta(\boldsymbol{z}_*))^2	\nonumber \\
		&=\check{r}_{\boldsymbol{z}_*}^{[i]}(t)^2+\check{e}_{\boldsymbol{z}_*}^{[i]}(t)^2+\eta(\boldsymbol{z}_*)^2+2\check{r}_{\boldsymbol{z}_*}^{[i]}(t)\check{e}_{\boldsymbol{z}_*}^{[i]}(t)\\
		&
		-2\check{r}_{\boldsymbol{z}_*}^{[i]}(t)\eta(\boldsymbol{z}_*)-2\check{e}_{\boldsymbol{z}_*}^{[i]}(t)\eta(\boldsymbol{z}_*).
	\end{align*}
	Taking the expectation on both sides of the above equation yields
		\begin{align}\label{eq:30}
		&\mathbb{E}[(\check{{\mu}}_{\boldsymbol{z}_*|\mathcal{D}^{[i]}(t)}-\eta(\boldsymbol{z}_*))^2] \nonumber \\
		=&\mathbb{E}[\check{r}_{\boldsymbol{z}_*}^{[i]}(t)^2]+\mathbb{E}[\check{e}_{\boldsymbol{z}_*}^{[i]}(t)^2]+\mathbb{E}[\eta(\boldsymbol{z}_*)^2]+2\mathbb{E}[\check{r}_{\boldsymbol{z}_*}^{[i]}(t)\check{e}_{\boldsymbol{z}_*}^{[i]}(t)]\nonumber \\
		&-2\mathbb{E}[\check{r}_{\boldsymbol{z}_*}^{[i]}(t)\eta(\boldsymbol{z}_*)]-2\mathbb{E}[\check{e}_{\boldsymbol{z}_*}^{[i]}(t)\eta(\boldsymbol{z}_*)].
		\end{align}

		By Lemma \ref{Lem:66}, plugging all terms into (\ref{eq:30}) renders
		\begin{align*}
		&\mathbb{E}[(\check{{\mu}}_{\boldsymbol{z}_*|\mathcal{D}^{[i]}(t)}-\eta(\boldsymbol{z}_*))^2] \nonumber \\
		=&\left(\frac{\kappa(D_{\boldsymbol{z}_*}^{\mathcal{Z}^{[i]}(t)})}{\sigma_f^2+(\sigma_e^{[i]})^2}\eta(\boldsymbol{z}_*^{[i]}(t))\right)^2+\left(\frac{\sigma_e^{[i]}\kappa(D_{\boldsymbol{z}_*}^{\mathcal{Z}^{[i]}(t)})}{\sigma_f^2+(\sigma_e^{[i]})^2}\right)^2 \\
		&+\eta(\boldsymbol{z}_*)^2-2\frac{\kappa(D_{\boldsymbol{z}_*}^{\mathcal{Z}^{[i]}(t)})}{\sigma_f^2+(\sigma_e^{[i]})^2}\eta(\boldsymbol{z}_*^{[i]}(t))\eta(\boldsymbol{z}_*) \\
		=& \left(\frac{\kappa(D_{\boldsymbol{z}_*}^{\mathcal{Z}^{[i]}(t)})}{\sigma_f^2+(\sigma_e^{[i]})^2}\eta(\boldsymbol{z}_*^{[i]}(t))-\eta(\boldsymbol{z}_*)\right)^2\nonumber \\
		&+\left(\frac{\sigma_e^{[i]}\kappa(D_{\boldsymbol{z}_*}^{\mathcal{Z}^{[i]}(t)})}{\sigma_f^2+(\sigma_e^{[i]})^2}\right)^2.
		\end{align*}		
		Since $\left(\frac{\kappa(D_{\boldsymbol{z}_*}^{\mathcal{Z}^{[i]}(t)})}{\sigma_f^2+(\sigma_e^{[i]})^2}\eta(\boldsymbol{z}_*^{[i]}(t))-\eta(\boldsymbol{z}_*)\right)^2\ge0$, then the mean squared error is lower bounded as
		\begin{align*}
		\mathbb{E}[(\check{{\mu}}_{\boldsymbol{z}_*|\mathcal{D}^{[i]}(t)}-\eta(\boldsymbol{z}_*))^2]\ge\left(\frac{\sigma_e^{[i]}\kappa(D_{\boldsymbol{z}_*}^{\mathcal{Z}^{[i]}(t)})}{\sigma_f^2+(\sigma_e^{[i]})^2}\right)^2.
		\end{align*}	
		Monotonicity of $\kappa(\cdot)$ in Assumption \ref{Asu:2} together with $d^{[i]}(t)\ge D_{\boldsymbol{z}_*}^{\mathcal{Z}^{[i]}(t)}$ renders $\kappa(D_{\boldsymbol{z}_*}^{\mathcal{Z}^{[i]}(t)})\ge\kappa(d^{[i]}(t))$, which implies $\mathbb{E}[(\check{{\mu}}_{\boldsymbol{z}_*|\mathcal{D}^{[i]}(t)}-\eta(\boldsymbol{z}_*))^2]\ge\left(\frac{\sigma_e^{[i]}\kappa(d^{[i]}(t))}{\sigma_f^2+(\sigma_e^{[i]})^2}\right)^2$.
		$\hfill\blacksquare$

		The following lemma characterizes the upper bound of the mean squared error for the predictive mean from the cloud-based aggregated GPR.
		
		\begin{Lem}\label{Lem:7}
			Suppose Assumptions \ref{Asu:00}, \ref{Asu:2} and \ref{Asu:3} hold. For all $\boldsymbol{z}_*\in{\mathcal{Z}_*}$, it holds that $\mathbb{E}[(\hat{{\mu}}_{\boldsymbol{z}_*|\mathcal{D}(t)}-\eta(\boldsymbol{z}_*))^2]\le\left(\Theta(d^{\max}(t))\right)^2$.
		\end{Lem}
		
		\emph{Proof:} Recall from definition (\ref{eq:7}) that $\hat{{\mu}}_{\boldsymbol{z}_*|\mathcal{D}(t)}= \frac{\hat{\sigma}^2_{\boldsymbol{z}_*|\mathcal{D}(t)}}{|\mathcal{I}(t)|}\sum_{i\in\mathcal{I}(t)}\check{{\mu}}'_{\boldsymbol{z}_{*}|\mathcal{D}^{[i]}(t)}\check{\sigma}'^{-2}_{\boldsymbol{z}_{*}|\mathcal{D}^{[i]}(t)}$.
		Then we have 
		\begin{align}
		&\left|\hat{{\mu}}_{\boldsymbol{z}_*|\mathcal{D}(t)}-\eta(\boldsymbol{z}_*)\right|\nonumber \\
		&=\left|\frac{\hat{\sigma}^2_{\boldsymbol{z}_*|\mathcal{D}(t)}}{|\mathcal{I}(t)|}\sum_{i\in\mathcal{I}(t)}\check{{\mu}}'_{\boldsymbol{z}_{*}|\mathcal{D}^{[i]}(t)}\check{\sigma}'^{-2}_{\boldsymbol{z}_{*}|\mathcal{D}^{[i]}(t)}-\eta(\boldsymbol{z}_*)\right| \nonumber \\
		&=\left|\frac{\hat{\sigma}^2_{\boldsymbol{z}_*|\mathcal{D}(t)}}{|\mathcal{I}(t)|}\sum_{i\in\mathcal{I}(t)}\check{{\mu}}'_{\boldsymbol{z}_{*}|\mathcal{D}^{[i]}(t)}\check{\sigma}'^{-2}_{\boldsymbol{z}_{*}|\mathcal{D}^{[i]}(t)} \right.\nonumber \\
		& \quad\left.-\frac{\hat{\sigma}^2_{\boldsymbol{z}_*|\mathcal{D}(t)}}{|\mathcal{I}(t)|}\sum_{i\in\mathcal{I}(t)}\eta(\boldsymbol{z}_*)\check{\sigma}'^{-2}_{\boldsymbol{z}_{*}|\mathcal{D}^{[i]}(t)}\right| \nonumber \\
		&\le\frac{\hat{\sigma}^2_{\boldsymbol{z}_*|\mathcal{D}(t)}}{|\mathcal{I}(t)|}\sum_{i\in\mathcal{I}(t)}\check{\sigma}'^{-2}_{\boldsymbol{z}_{*}|\mathcal{D}^{[i]}(t)}\left|\check{{\mu}}'_{\boldsymbol{z}_{*}|\mathcal{D}^{[i]}(t)}-\eta(\boldsymbol{z}_*)\right|.\label{eq:300}
		\end{align}
		{{By Lemma \ref{Lem:55}, we have $\left|\check{{\mu}}'_{\boldsymbol{z}_{*}|\mathcal{D}^{[i]}(t)}-\eta(\boldsymbol{z}_*)\right|\le\Theta(d^{\max}(t))$.}}
		Since $\frac{\hat{\sigma}^2_{\boldsymbol{z}_*|\mathcal{D}(t)}}{|\mathcal{I}(t)|}\sum_{i\in\mathcal{I}(t)}\check{\sigma}'^{-2}_{\boldsymbol{z}_{*}|\mathcal{D}^{[i]}(t)}=1$, then plugging {{the upper bound $\Theta(d^{\max}(t))$}} into (\ref{eq:300}) renders 
		\begin{align*}
		&\left|{{\hat{{\mu}}_{\boldsymbol{z}_*|\mathcal{D}(t)}}}-\eta(\boldsymbol{z}_*)\right|\le\Theta(d^{\max}(t)),
		\end{align*}
		which implies $\mathbb{E}[(\hat{{\mu}}_{\boldsymbol{z}_*|\mathcal{D}(t)}-\eta(\boldsymbol{z}_*))^2]\le\left(\Theta(d^{\max}(t))\right)^2$.
		$\hfill\blacksquare$
		
		
		{{First of all, according to Theorem IV.3 in \cite{ZM2021}, we have $\check{\sigma}_{\boldsymbol{z}_*|\mathcal{D}^{[i]}(t)}^2 \le \sigma_f^2-\frac{\kappa(d^{[i]}(t))^2}{\sigma_f^2+(\sigma_e^{[i]})^2}$ for all $\boldsymbol{z}_*\in\mathcal{Z}_*$. Second, recall that $\check{\theta}^{[i]}(t)=\frac{\sigma_e^{[i]}\kappa(d^{[i]}(t))}{\sigma_f^4+\sigma_f^2(\sigma_e^{[i]})^2-\kappa(d^{[i]}(t))^2}$. Then, we have $\check{\sigma}_{\boldsymbol{z}_*|\mathcal{D}^{[i]}(t)}^2\check{\theta}^{[i]}(t)\leq\frac{\sigma_e^{[i]}\kappa(d^{[i]}(t))}{\sigma_f^2+(\sigma_e^{[i]})^2}$. Since $\check{\theta}^{[i]}(t)>0$ and $\check{\sigma}_{\boldsymbol{z}_*|\mathcal{D}^{[i]}(t)}^2>0$}}, by Lemma \ref{Lem:6} and monotonicity of $\kappa(\cdot)$ in Assumption \ref{Asu:2}, we have
		\begin{align}\label{eq:277}
			\mathbb{E}[(\check{{\mu}}_{\boldsymbol{z}_*|\mathcal{D}^{[i]}(t)}-\eta(\boldsymbol{z}_*))^2]&\ge\left(\frac{\sigma_e^{[i]}\kappa(d^{[i]}(t))}{\sigma_f^2+(\sigma_e^{[i]})^2}\right)^2\nonumber \\&\ge\left(\check{{\sigma}}^2_{\boldsymbol{z}_*|\mathcal{D}^{[i]}(t)}\check{\theta}^{[i]}(t)\right)^2.
		\end{align}
		Recall that $\Theta(d^{\max}(t))=(1-\frac{\kappa(d^{\max}(t))}{\sigma_f^2+(\sigma_e^{\max})^2})\|\eta\|_{\infty}+\frac{\sigma_f^2\ell_{\eta} d^{\max}(t)}{\sigma_f^2+(\sigma_e^{\min})^2}+\sqrt{2\sigma^2(\ln2-\ln\delta)}$. We would like to find an upper bound of the first term on the right-hand side of $\Theta(d^{\max}(t))$, i.e., $\frac{\sigma_f^2+(\sigma_e^{\max})^2-\kappa(d^{\max}(t))}{\sigma_f^2+(\sigma_e^{\max})^2}\|\eta\|_{\infty}$. By Part II of Theorem \ref{Thm:2}, we have ${{(1-2\beta)}}\left(\frac{\sigma_f^2(\sigma_e^{\min})^2}{\sigma_f^2+(\sigma_e^{\max})^2}\right)\le\hat{\sigma}^2_{\boldsymbol{z}_*|\mathcal{D}(t)}$, {{which indicates $\frac{1}{\sigma_f^2+(\sigma_e^{\max})^2}\le\frac{\hat{\sigma}^2_{\boldsymbol{z}_*|\mathcal{D}(t)}}{{{(1-2\beta)}}\sigma_f^2(\sigma_e^{\min})^2}$}}. Then, the first term {{$(1-\frac{\kappa(d^{\max}(t))}{\sigma_f^2+(\sigma_e^{\max})^2})\|\eta\|_{\infty}$ on the right-hand side of $\Theta(d^{\max}(t))$}} is upper bounded by
		\begin{align*}
			&(1-\frac{\kappa(d^{\max}(t))}{\sigma_f^2+(\sigma_e^{\max})^2})\|\eta\|_{\infty} \nonumber \\
			&\le\frac{\sigma_f^2+(\sigma_e^{\max})^2-\kappa(d^{\max}(t))}{{{(1-2\beta)}}\sigma_f^2(\sigma_e^{\min})^2}\|\eta\|_{\infty}\hat{\sigma}^2_{\boldsymbol{z}_*|\mathcal{D}(t)}.
		\end{align*}
		Recall that $\hat{\theta}=\frac{\sigma_f^2+(\sigma_e^{\max})^2-\kappa(\gamma_d)}{{{(1-2\beta)}}\sigma_f^2(\sigma_e^{\min})^2}\|\eta\|_{\infty}$, $\Gamma = \frac{\sigma_f^2\ell_{\eta} \gamma_d}{\sigma_f^2+(\sigma_e^{\min})^2}+\sqrt{2\sigma^2(\ln2-\ln\delta)}$ and $d^{[i]}(t)\le\gamma_d$ for all $i\in\mathcal{V}$. It follows from Lemma \ref{Lem:7} that
		\begin{align}\label{eq:33}
		&\mathbb{E}[(\hat{{\mu}}_{\boldsymbol{z}_*|\mathcal{D}(t)}-\eta(\boldsymbol{z}_*))^2]\nonumber \\
		&\le\left(\frac{\sigma_f^2+(\sigma_e^{\max})^2-\kappa(d^{\max}(t))}{{{(1-2\beta)}}\sigma_f^2(\sigma_e^{\min})^2}\|\eta\|_{\infty}\hat{\sigma}^2_{\boldsymbol{z}_*|\mathcal{D}(t)}\right. \nonumber \\
		& \left.\quad+\frac{\sigma_f^2\ell_{\eta} d^{\max}(t)}{\sigma_f^2+(\sigma_e^{\min})^2}+\sqrt{2\sigma^2(\ln2-\ln\delta)}\right)^2 \nonumber \\
		& \le\left(\hat{\theta}\hat{\sigma}^2_{\boldsymbol{z}_*|\mathcal{D}(t)}+\Gamma\right)^2.
		\end{align}
		If $\boldsymbol{z}_*\in\mathcal{Z}_{fused}^{[i]}(t)$, then for  $i\in\mathcal{V}\backslash\mathcal{B}$, we have $\hat{\theta}\hat{\sigma}^2_{\boldsymbol{z}_*|\mathcal{D}(t)}+\Gamma<\check{\theta}^{[i]}(t)\check{{\sigma}}^2_{\boldsymbol{z}_*|\mathcal{D}^{[i]}(t)}$. Comparing (\ref{eq:277}) and (\ref{eq:33}) further renders $\mathbb{E}[(\tilde{\mu}^{[i]}_{\boldsymbol{z}_{*}|\mathcal{D}(t)}-\eta(\boldsymbol{z}_*))^2-(\check{{\mu}}_{\boldsymbol{z}_*|\mathcal{D}^{[i]}(t)}-\eta(\boldsymbol{z}_*))^2]<0$.
		
		If $\boldsymbol{z}_*\notin\mathcal{Z}_{fused}^{[i]}(t)$, we have $\tilde{{\mu}}^{[i]}_{\boldsymbol{z}_*|\mathcal{D}(t)}=\check{\mu}_{\boldsymbol{z}_*|\mathcal{D}^{[i]}(t)}$, which gives $\mathbb{E}[(\tilde{\mu}^{[i]}_{\boldsymbol{z}_{*}|\mathcal{D}(t)}-\eta(\boldsymbol{z}_*))^2-(\check{{\mu}}_{\boldsymbol{z}_*|\mathcal{D}^{[i]}(t)}-\eta(\boldsymbol{z}_*))^2] =0$.
		
		\textbf{Part III:} Pick any $i\in\mathcal{V}\backslash\mathcal{B}$. If $\boldsymbol{z}_*\in\mathcal{Z}_{fused}^{[i]}(t)$, then we have $(\tilde{\sigma}^{[i]}_{\boldsymbol{z}_*|\mathcal{D}(t)})^{2}=\hat{\sigma}^2_{\boldsymbol{z}_{*}|\mathcal{D}(t)}$. By Part II of Theorem \ref{Thm:2}, we have ${{(1-2\beta)}}\left(\frac{\sigma_f^2(\sigma_e^{\min})^2}{\sigma_f^2+(\sigma_e^{\max})^2}\right)\le(\tilde{\sigma}^{[i]}_{\boldsymbol{z}_*|\mathcal{D}(t)})^{2}\le {{\frac{1}{1-4\beta}}}\left(\sigma_f^2-\frac{\kappa(d^{\max}(t))^2}{\sigma_f^2+(\sigma_e^{\max})^2}\right)$.
		
		If $\boldsymbol{z}_*\notin\mathcal{Z}_{fused}^{[i]}(t)$, then we have $(\tilde{\sigma}^{[i]}_{\boldsymbol{z}_*|\mathcal{D}(t)})^{2}=\check{\sigma}^2_{\boldsymbol{z}_{*}|\mathcal{D}^{[i]}(t)}$. By Theorem IV.3 in \cite{ZM2021}, we have $\frac{\sigma_f^2(\sigma_e^{[i]})^2}{\sigma_f^2+(\sigma_e^{[i]})^2}\le(\tilde{\sigma}^{[i]}_{\boldsymbol{z}_*|\mathcal{D}(t)})^{2} \le \sigma_f^2-\frac{\kappa(d^{[i]}(t))^2}{\sigma_f^2+(\sigma_e^{[i]})^2}$. Monotonicity of $\kappa(\cdot)$ in Assumption \ref{Asu:2} gives $\kappa(d^{\max}(t))\le\kappa(d^{[i]}(t))$, and Assumption \ref{Asu:00} gives  $\frac{1}{1-4\beta}>1$ and $1-2\beta<1$. Therefore, the desired inequality holds for all $\boldsymbol{z}_*\in\boldsymbol{\mathcal{Z}}$.

\section{Numerical experiments}
This section experimentally demonstrates the performances of the proposed Byzantine-resilient federated GPR algorithm using a toy example and two medium-scale real-world datasets. We conduct the experiments on a computer with Intel $i7$-$6600$ CPU, $2.60$GHz and $12$ GB RAM. Specifically, the first two experiments use a synthetic dataset to evaluate the prediction performance in terms of consistency, and the effects of parameters $\alpha$ and $\beta$. Then the third and fourth experiments use the synthetic dataset to show the effectiveness of learning different functions and the robustness to different attacks, respectively. The fifth experiment evaluates the learning accuracy improvement of the agent-based fused GPR over the agent-based local GPR. The sixth experiment evaluates the learning performance by comparing with the state-of-the-art algorithms. We leverage two real-world datasets to evaluate the prediction performance on different attack magnitudes and learning accuracy improvements of the agent-based fused GPR, respectively.

\subsection{Toy example:} 
For the observation model (\ref{eq:2}), we consider the following target function introduced in \cite{HJYY2018} 
\begin{align}\label{eq:8}
\eta(\boldsymbol{z}) &=  (\boldsymbol{z}^3-0.5)\sin(3\boldsymbol{z}-0.5)\nonumber \\
&\quad+5\boldsymbol{z}^2\sin(12\boldsymbol{z}) +4\cos(2\boldsymbol{z}),
\end{align}
with $\boldsymbol{z}\in\mathbb{R}$. {{We add Gaussian noise $e\sim\mathcal{N}(0,0.01)$ into the target function $\eta(\boldsymbol{z})$ to construct the measurement $y$}}. We generate $n_s = 10^3,5\times10^3,10^4,5\times10^4$ training points in $[0,1]$, respectively, and choose $n_t = 120$ test points randomly in $[0,1]$. There are $n=40$ agents in a network. We partition the training dataset into $40$ disjoint groups, and each agent is assigned with $n_s/n$ training data points. We use the following squared-exponential kernel
$
k(\boldsymbol{z},\boldsymbol{z}_*) = \sigma_f^2\exp({-\frac{1}{2\ell^2}}(\boldsymbol{z}-\boldsymbol{z}_*)^2)$, and use Mean Squared Error (MSE) to evaluate the proposed algorithm in the following experiments.   

\begin{figure}[htbp]
	\centering
	\vspace{-0.5cm}
	\subfigure[Consistency evaluation]{
		\begin{minipage}[t]{0.5\linewidth}
			\centering
			\includegraphics[width=1.7in,height=1.5in]{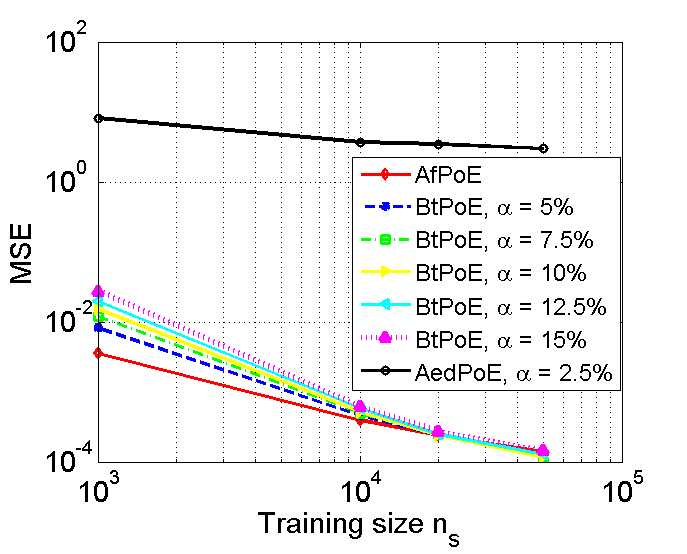}
			\label{Fig:1a}
		\end{minipage}%
	}%
	\subfigure[Prediction evaluation on $\beta$]{
		\begin{minipage}[t]{0.5\linewidth}
			\centering
			\includegraphics[width=1.7in,height=1.5in]{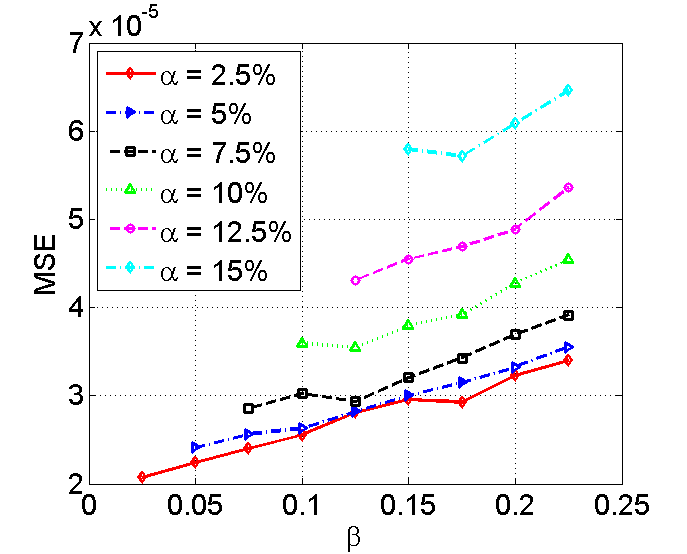}
			\label{Fig:1b}
		\end{minipage}%
	}%
	\centering
	\caption{Prediction performance of the cloud-based aggregated GPR.}\label{Fig:1}
\end{figure}

\subsubsection{Prediction performance in terms of consistency and different $\alpha$} In the first experiment, we let $\alpha=2.5\%,5\%,7.5\%,10\%$, $12.5\%,15\%$. We randomly choose the agents in the network to be compromised by same-value attacks  \citep{LLXW2019}, and let $\beta=\alpha$. Specifically, for each test point $\boldsymbol{z}_*$, the Byzantine agents only change the local predictive means to $100$, that is, $\check{{\mu}}'_{\boldsymbol{z}_{*}|\mathcal{D}^{[i]}(t)}=100$ for all $t$, and send this incorrect prediction to the cloud. The local predictive variances for all the agents remain unchanged. Fig. \ref{Fig:1a} compares the performances of the proposed Byzantine-resilient PoE (BtPoE) with the standard PoE. As a benchmark, the bottom curve depicts the prediction errors of the standard PoE (AfPoE) when there is no Byzantine attack in the network; it shows that the standard PoE is consistent in the attack-free scenario. However, when the aforementioned Byzantine attacks are launched, the top curve shows that the standard PoE can no longer remain consistent. Note that there are $6$ curves by attacked standard PoE (AedPoE), and only the closest one to the bottom curve is shown in Fig. \ref{Fig:1a}.

\begin{figure}[htbp]
	\vspace{-0.4cm}
	\subfigure[Attack-free standard PoE]{
		\begin{minipage}[t]{0.5\linewidth}
			\centering
			\includegraphics[width=1.7in,height=1.5in]{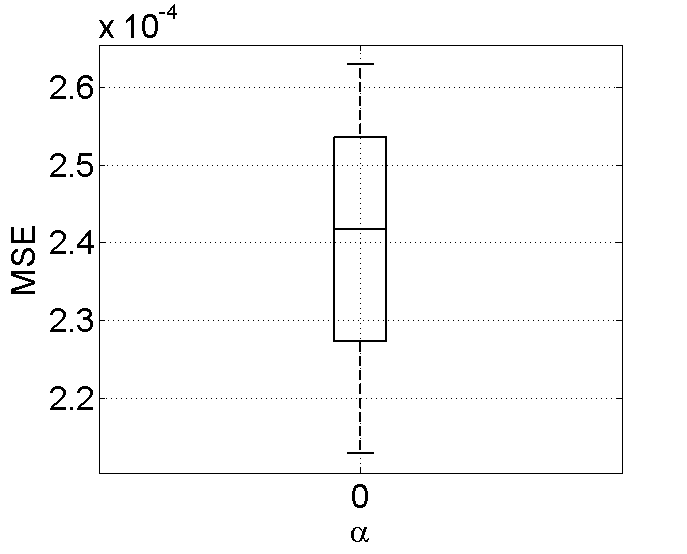}
			\label{Fig:11a}
		\end{minipage}%
	}%
	\subfigure[Byzantine-resilient PoE]{
		\begin{minipage}[t]{0.5\linewidth}
			\centering
			\includegraphics[width=1.7in,height=1.5in]{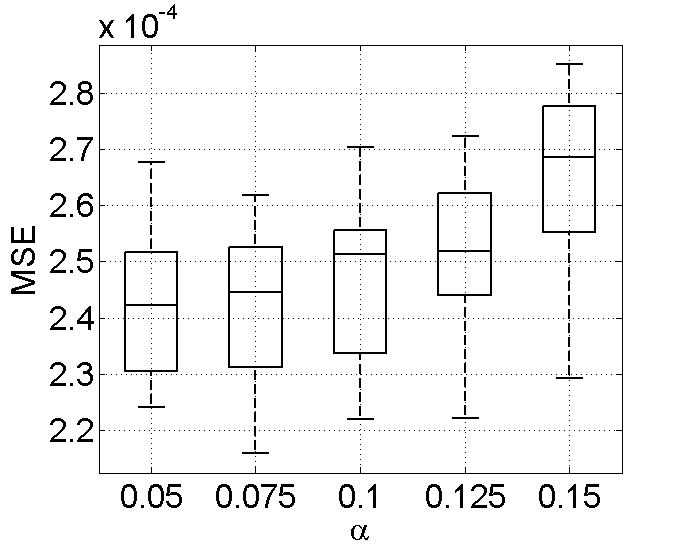}
			\label{Fig:11b}
		\end{minipage}%
	}%
	
	\subfigure[Attacked standard PoE]{
		\begin{minipage}[t]{0.5\linewidth}
			\centering
			\includegraphics[width=1.7in,height=1.5in]{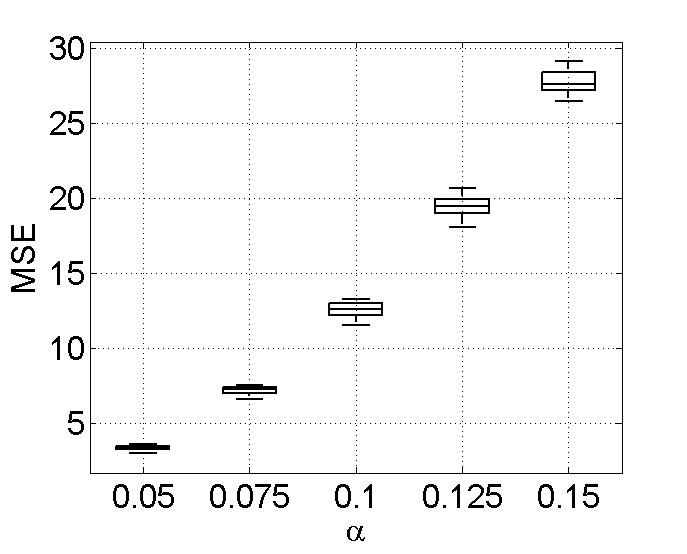}
			\label{Fig:11c}
		\end{minipage}%
	}%
	\centering
	\caption{Comparison of the prediction performance on $\alpha$.}\label{Fig:11}
\end{figure}

Notice that the performance of the Byzantine-resilient PoE increases as $\alpha$, the ratio of Byzantine agents, decreases. Fixing the training data size $n_s=10^4$, we replicate the experiments for $50$ times with different $\alpha$ and observation noise $e$. Figs. \ref{Fig:11a} and \ref{Fig:11b} show that the performances of the Byzantine-resilient PoE are comparable to that of the attack-free standard PoE with consistency preserved. The prediction errors for both methods are in the order of $10^{-4}$, while Fig. \ref{Fig:11c} shows that the attacked standard PoE has prediction errors ranged from $2.5$ to $30$. Therefore, the proposed PoE aggregation rule has strong ability to tolerate Byzantine attacks.

\subsubsection{Prediction performance on different $\beta$}In the second experiment, we evaluate the prediction performance with regard to $\beta$. All associated parameters are the same as those in the above experiments. Notice that for each $\alpha$, we initially let $\beta=\alpha$ and then increase $\beta$ to $0.225$ with increment $0.025$, such that the assumption $\alpha\le\beta<\frac{1}{4}$ is satisfied. The results are shown in Fig. \ref{Fig:1b}. This shows that as the number of trimmed agents increases, the performance of the proposed algorithm grows worse, since more information is excluded in the aggregation. This result validates the conclusion in Theorem \ref{Thm:2}.

\begin{table}[H]
	\caption{Comparison of the prediction performance on different functions. The order of MSE is $10^{-3}$. }
	\label{Tab:1}
	\begin{tabular}{|c|c|c|c|}
		\hline
		Algorithm & AfPoE  & BtPoE & AedPoE \\
		\hline
		MSE & $4.9\pm0.007$ & $23.6\pm0.172$ & $26533.9\pm0.019$ \\
		\hline
	\end{tabular}
\end{table}

\subsubsection{Prediction performance on different functions} In the third experiment, we conduct Monte Carlo simulations to evaluate the performance of Byzantine-resilient PoE on different functions. We repeat the experiment for $50$ times by sampling different functions and observation points. Specifically, we replace $\sin(12\boldsymbol{z})$ in (\ref{eq:8}) with $\sin(12\boldsymbol{z})+\epsilon'$ where $\epsilon'\sim\mathcal{N}(0,0.01i)$, $i=1,\ldots,50$. We assign $n=40$ agents in the network and sample $n_s = 10^4$ training points in $[0,1]$, and then choose $n_t = 120$ test points randomly in $[0,1]$. We also use the square-exponential kernel $k(\boldsymbol{z},\boldsymbol{z}_*)$. There are six Byzantine agents, i.e., $\alpha=15\%$, and we let $\beta=\alpha$. Experiment results are presented in Table \ref{Tab:1}, and it shows that the performance of the Byzantine-resilient PoE is comparable to that of the attack-free standard PoE, noticing that the order of the prediction errors for the Byzantine-resilient PoE is $10^{-2}$. For comparison, the MSE by the attacked standard PoE is about $26.534$. This demonstrates that the proposed Byzantine-resilient aggregation rule is effective for learning different functions.


\begin{figure}[htbp]
	\centering
	
	\subfigure[Under Gaussian attacks.]{
		\begin{minipage}[t]{0.5\linewidth}
			\centering
			\includegraphics[width=1.6in,height=1.5in]{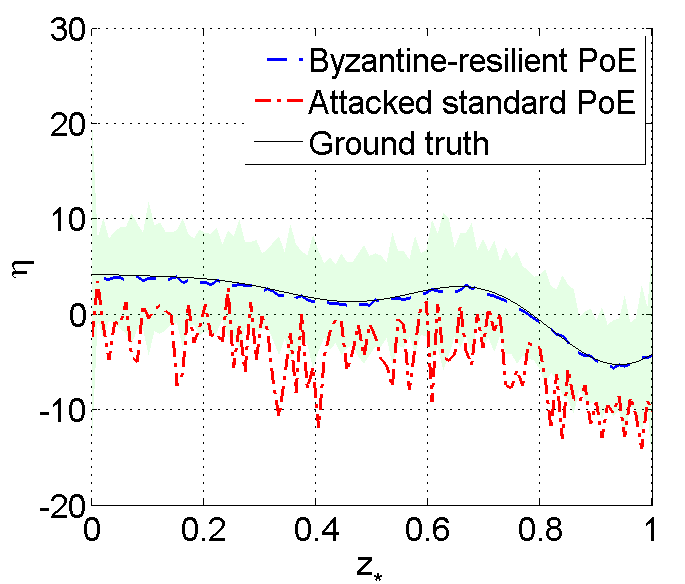}
			\label{Fig:0a}
		\end{minipage}%
	}%
	\subfigure[Under ALTE attacks.]{
		\begin{minipage}[t]{0.5\linewidth}
			\centering
			\includegraphics[width=1.6in,height=1.5in]{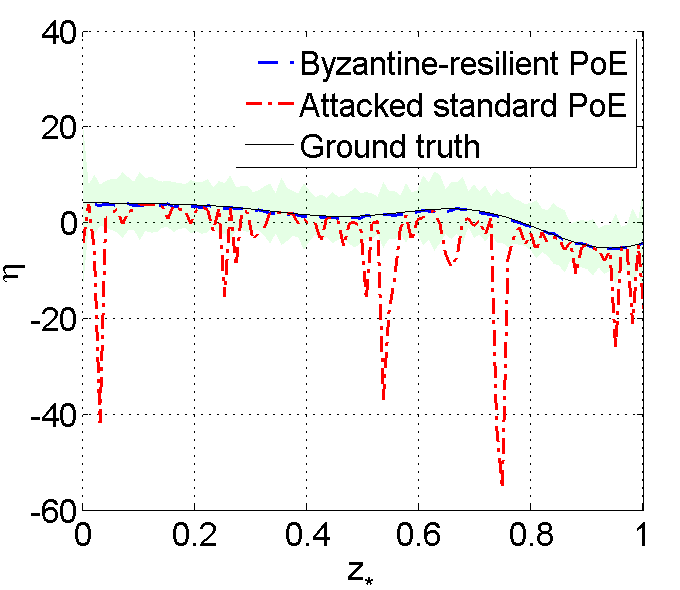}
			\label{Fig:0b}
		\end{minipage}%
	}%
	
	\subfigure[Under mimic attacks.]{
		\begin{minipage}[t]{0.5\linewidth}
			\centering
			\includegraphics[width=1.6in,height=1.5in]{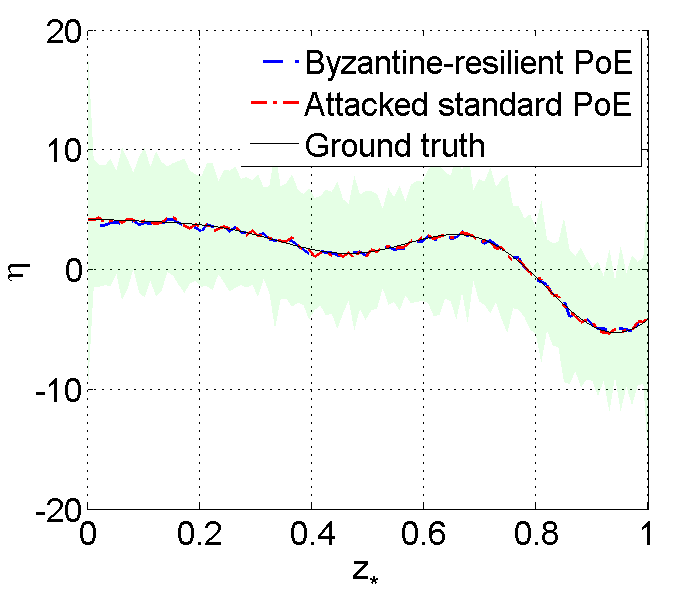}
			\label{Fig:0c}
		\end{minipage}%
	}%
	\subfigure[Under bit-flipped attacks.]{
		\begin{minipage}[t]{0.5\linewidth}
			\centering
			\includegraphics[width=1.6in,height=1.5in]{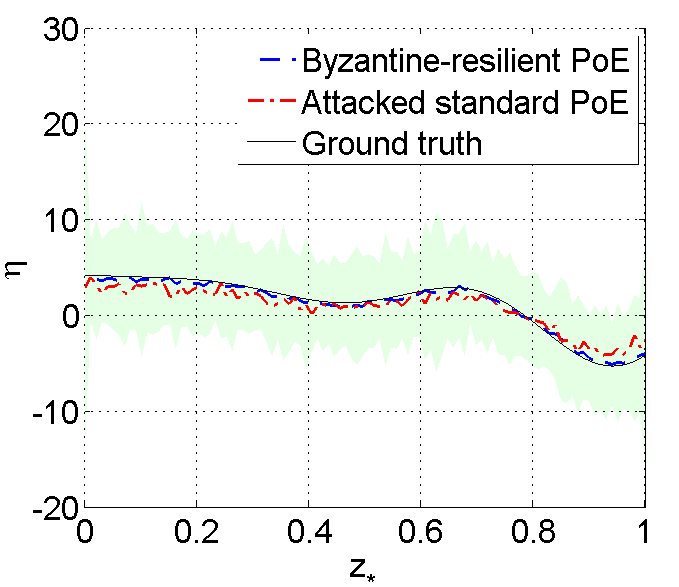}
			\label{Fig:0d}
		\end{minipage}%
	}%
	
	\centering
	\caption{Prediction comparisons of the cloud-based aggregated GPR under $4$ attacks.}\label{Fig:00}
\end{figure}
\subsubsection{Prediction performances on different attacks}

In the fourth experiment, we evaluate the robustness of the Byzantine-resilient PoE to $4$ types of Byzantine attacks. Following \cite{SPLM2022}, \cite{ESP2022}, \cite{allouah2023fixing} and \cite{LLXW2019}, we choose Gaussian, ALTE, mimic and bit-flipped attacks in our experiments. We fix $n_s=4\times10^{4}$ and $\alpha=\beta=15\%$. The prediction performances are given in Fig. \ref{Fig:00}, which shows that the learning performances of the Byzantine-resilient PoE are comparable to those of the attack-free standard PoE with consistency preserved. Therefore, our proposed algorithm is demonstrated to be resilient to Byzantine attacks in the network.

\begin{figure}[htbp]
	\centering
	\subfigure[Avergaed prediction errors]{
		\begin{minipage}[t]{0.5\linewidth}
			\centering
			\includegraphics[width=1.6in,height=1.5in]{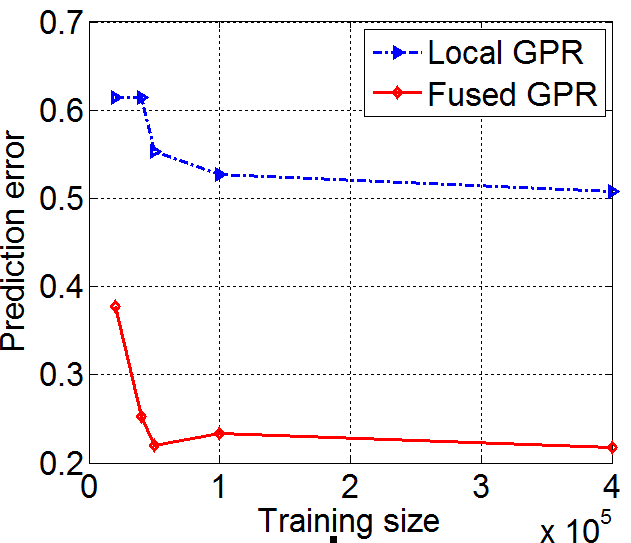}
			\label{Fig:2a}
		\end{minipage}%
	}%
	\subfigure[Averaged variances]{
		\begin{minipage}[t]{0.5\linewidth}
			\centering
			\includegraphics[width=1.6in,height=1.5in]{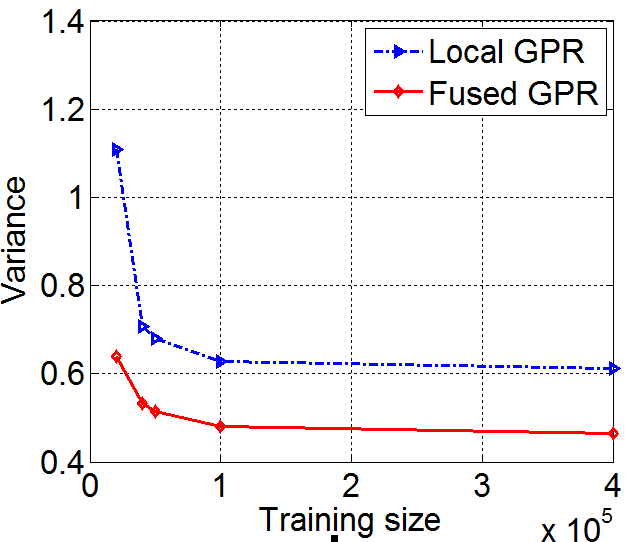}
			\label{Fig:2b}
		\end{minipage}%
	}%
	\centering
	\caption{Prediction performance comparison of the agent-based local GPR and fused GPR on the toy example.}\label{Fig:2}
\end{figure}

\subsubsection{Agent-based fused GPR versus agent-based local GPR} In the fifth experiment, we evaluate the learning performance improvements of the agent-based fused GPR over the agent-based local GPR by using (\ref{eq:88}). We also assign $n=40$ agents in the network and choose $n_t = 120$ test points randomly in $[0,1]$. We also use the square-exponential kernel $k(\boldsymbol{z},\boldsymbol{z}_*)$. There are six Byzantine agents, i.e., $\alpha=15\%$, and we let $\beta=\alpha$. In $\mathcal{I}(t)$, agent $2$ is chosen and for different training data size, its prediction performances of the agent-based local GPR and fused GPR are shown in Fig. \ref{Fig:2}. Specifically, Fig. \ref{Fig:2}(a) depicts the comparison of averaged prediction errors for all test points between the agent-based local GPR and the agent-based fused GPR. The smaller prediction error of the agent-based fused GPR demonstrates the learning improvements by leveraging communications. Fig. \ref{Fig:2}(b) gives the comparison of averaged variances for all test points between the agent-based local GPR and fused GPR, which shows that the agent-based fused GPR can provide a smaller confidence interval for the predictions. 

\begin{table}[H]
	\caption{Agents 33 and 40's prediction performance comparisons between local GPR and fused GPR  {{on synthetic datasets}}.}
	\label{Tab:2}
	\centering
	\begin{tabular}{|c|c|c|c|c|}
		\hline
		Training data size & $6000$ & $35000$ & $50000$ & $200000$ \\
		\hline
		MSE/$i=33$/local & $1.1019$ & $1.1250$ & $1.0887$ & $1.0595$ \\
		\hline
		MSE/$i=33$/fused & $0.4771$ & $0.4954$ & $0.4855$ & $0.4444$ \\
		\hline
		MSE/$i=40$/local & $1.0818$ & $1.0656$ & $1.0725$ & $1.0534$ \\
		\hline
		MSE/$i=40$/fused & $0.2382$ & $0.2376$ & $0.2352$ & $0.2285$ \\
		\hline
		{{MSE/{{cloud}}}} & {{$0.6492$}} & {{$0.6394$}} & {{$0.6567$}} & {{$0.6191$}} \\
		\hline
	\end{tabular}
\end{table}
Moreover, the prediction performance comparisons between the agent-based local GPR and fused GPR on agents {{33 and 40}} are listed in Table \ref{Tab:2}. For a test point, the local variances $\check{\sigma}'^2_{\boldsymbol{z}_*|\mathcal{D}^{[i]}(t)}$ lie in $[0.3660,0.6228]$, and the global variance is $\hat{\sigma}^2_{\boldsymbol{z}_*|\mathcal{D}(t)}=0.4678$. Table \ref{Tab:2} shows that the learning performance of the agent-based fused GPR is better than that of the agent-based local GPR. This demonstrates that the comparison between variances in step 2 and step 7 in Algorithm \ref{Alg:4} is able to improve the learning accuracy, and our algorithm is effective. 

{{
\begin{figure}[H]
	\centering
	\vspace{-0.2cm}
	\subfigure[ Byzantine-resilient PoE versus median]{
		\begin{minipage}[t]{0.5\linewidth}
			\centering
			\includegraphics[width=1.6in,height=1.5in]{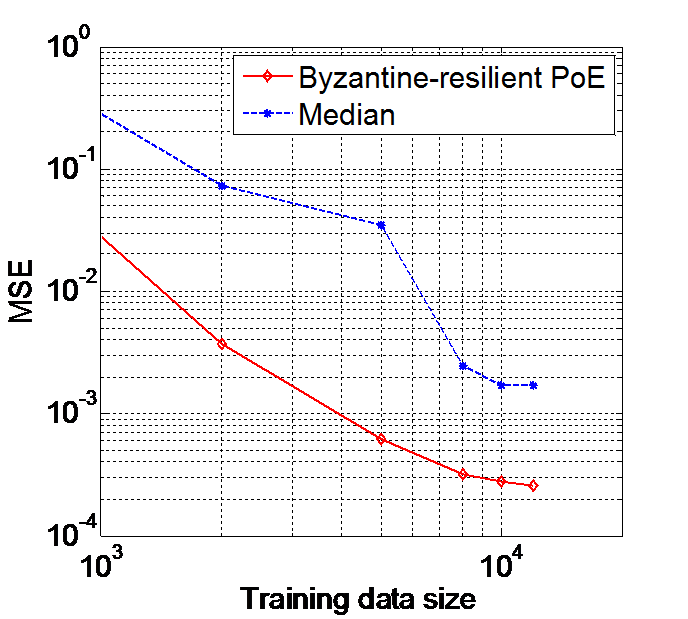}
			\label{Fig:111a}
		\end{minipage}%
	}%
	\subfigure[PoE versus average]{
		\begin{minipage}[t]{0.5\linewidth}
			\centering
			\includegraphics[width=1.6in,height=1.5in]{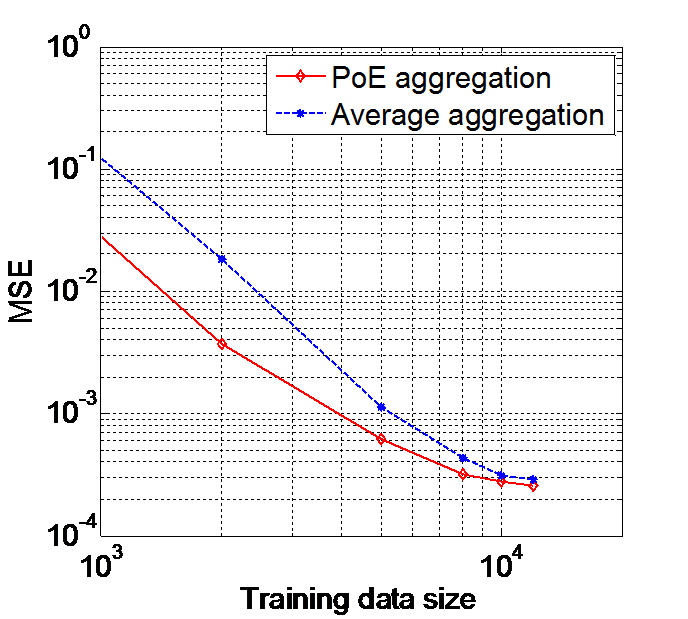}
			\label{Fig:111b}
		\end{minipage}%
	}%
	\centering
	\caption{Prediction performance comparisons between the proposed algorithm and the state-of-the-art algorithms.}\label{Fig:111}
\end{figure}
\subsubsection{Prediction performance comparisons between Byzantine-resilient PoE and the state-of-the-art algorithms}

In the sixth experiment, we first evaluate the learning performance by comparing with  median in \cite{YCRB2018}. We use the same experiment setup as Section 6.1.4, and the prediction performance of the cloud-based aggregated GPR is given in Fig. \ref{Fig:111a}. It can be seen that for each training data size, the prediction performance of Byzantine-resilient PoE is better than that of median. This is because Byzantine-resilient PoE leverages more data for predictions than median, and more local data can increase the learning performance. Then we compare PoE aggregation algorithm with average aggregation  \citep{KJM2016, MBM2017} in deep learning. The prediction performance of the PoE aggregation in Fig. \ref{Fig:111b} is superior to that of the average aggregation for each training data size.}}

\begin{figure}[H]
	\centering
	\vspace{-0.2cm}
	\subfigure[Attack-free standard PoE]{
		\begin{minipage}[t]{0.5\linewidth}
			\centering
			\includegraphics[width=1.7in,height=1.5in]{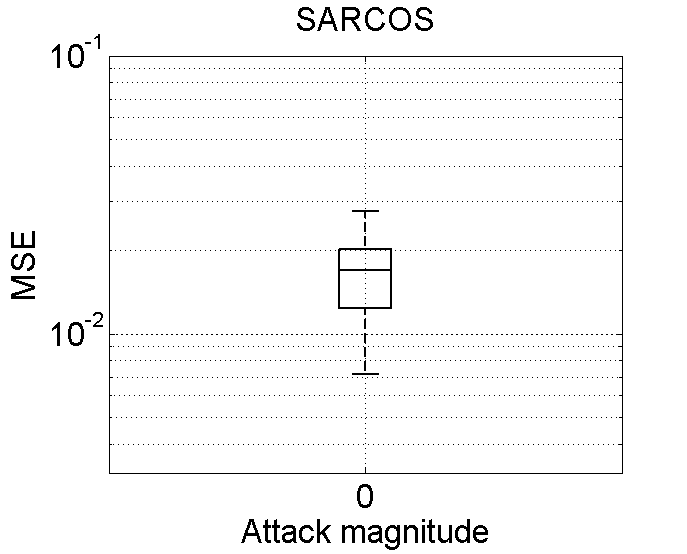}
			\label{Fig:3a}
		\end{minipage}%
	}%
	\subfigure[Byzantine-resilient PoE]{
		\begin{minipage}[t]{0.5\linewidth}
			\centering
			\includegraphics[width=1.7in,height=1.5in]{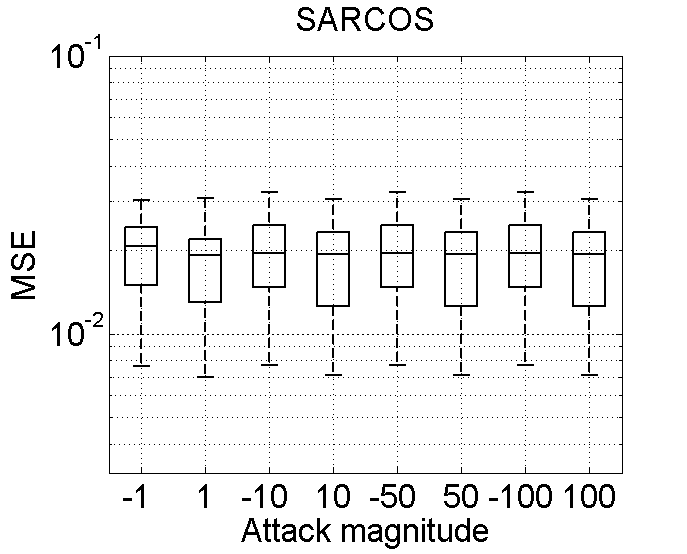}
			\label{Fig:3b}
		\end{minipage}%
	}%
	
	\subfigure[Attacked standard PoE]{
		\begin{minipage}[t]{0.5\linewidth}
			\centering
			\includegraphics[width=1.7in,height=1.5in]{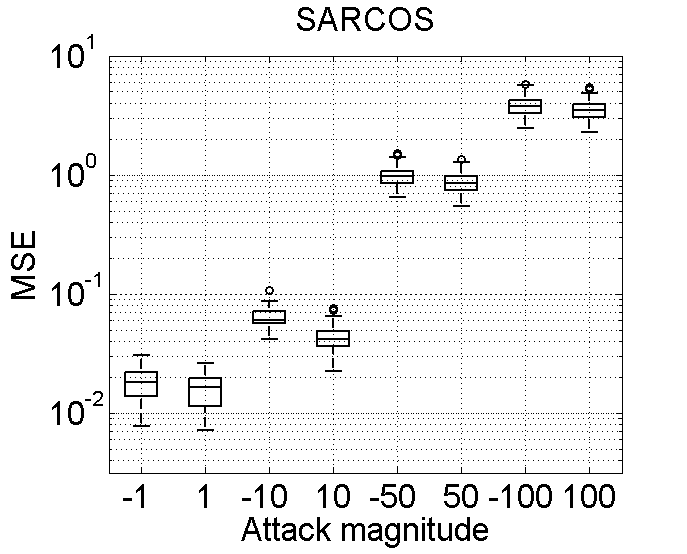}
			\label{Fig:3c}
		\end{minipage}%
	}%
	\subfigure[Attack-free standard PoE]{
		\begin{minipage}[t]{0.5\linewidth}
			\centering
			\includegraphics[width=1.7in,height=1.5in]{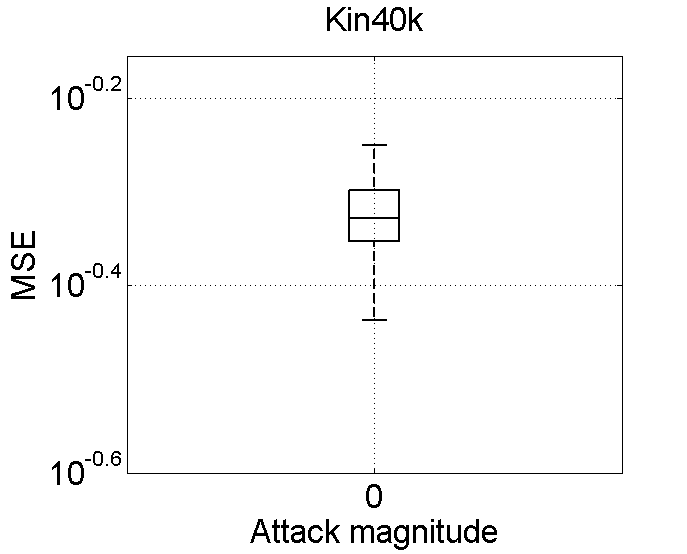}
			\label{Fig:3d}
		\end{minipage}%
	}%
	
	\subfigure[Byzantine-resilient PoE]{
		\begin{minipage}[t]{0.5\linewidth}
			\centering
			\includegraphics[width=1.7in,height=1.5in]{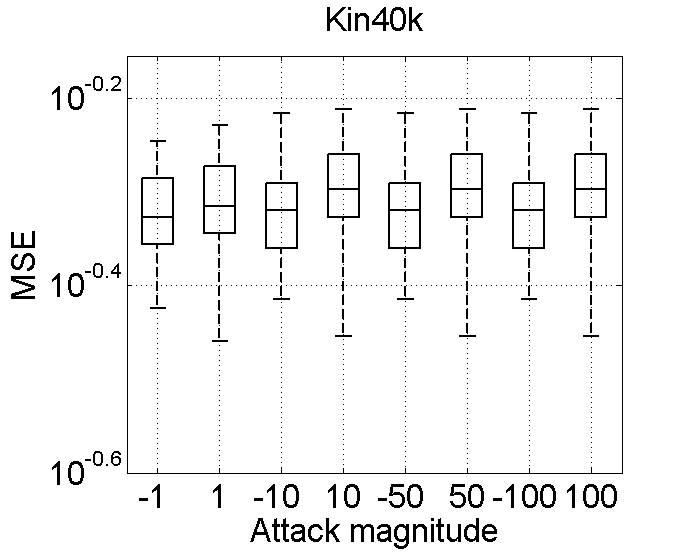}
			\label{Fig:3e}
		\end{minipage}%
	}%
	\subfigure[Attacked standard PoE]{
		\begin{minipage}[t]{0.5\linewidth}
			\centering
			\includegraphics[width=1.7in,height=1.5in]{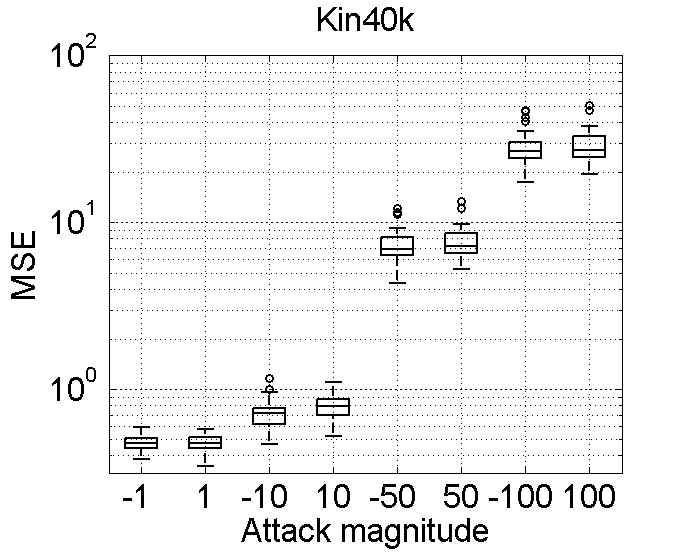}
			\label{Fig:3f}
		\end{minipage}%
	}%
	\centering
	\caption{Prediction performance with different attack magnitudes on datasets SARCOS and \emph{Kin40k}.}\label{Fig:3}
\end{figure}

\subsection{Medium-scale real-world datasets:}
The first dataset is collected from a seven degrees-of-freedom SARCOS anthropomorphic robot arm  \citep{CC2006}. The goal is to learn a function $\eta$. The input of $\eta$ is $21$-dimensional including $7$ joint positions, $7$ joint velocities and $7$ joint accelerations; whereas the output is $1$-dimensional, representing the torque of a joint. We randomly sample 40,000 data points for training and 40 data points for testing. The second dataset \emph{Kin40k} \citep{SWL2003} is created using a robot arm simulator. We randomly choose $9,000$ training points and $40$ test points. 

\subsubsection{Performance evaluation on different attack magnitudes} We conduct an experiment to evaluate the prediction performance with different attack magnitudes using the above two real-world datasets. The magnitudes of attack we consider are $\pm1,\pm10,\pm50,\pm100$. In the experiment, we partition training data evenly into $n=100$ disjoint groups, and assign each group to an agent. We randomly sample the training data points and repeat the experiments for $50$ times. Fig. \ref{Fig:3} shows the prediction performances. For both real-world datasets, when the attack magnitude is increasing, the prediction errors of the attacked standard PoE largely increases, whereas those of the attack-free standard PoE and the Byzantine-resilient PoE are consistently maintained at a low level. The proposed Byzantine-resilient PoE is resilient to the attack.

\subsubsection{Agent-based fused GPR versus local GPR}

We use datasets SARCOS and \emph{Kin40k} to evaluate the learning performance improvements of the agent-based fused GPR over the agent-based local GPR. 

\begin{figure}[H]
	\centering
	\vspace{-0.2cm}
	\subfigure[Sarcos]{
		\begin{minipage}[t]{0.5\linewidth}
			\centering
			\includegraphics[width=1.7in,height=1.5in]{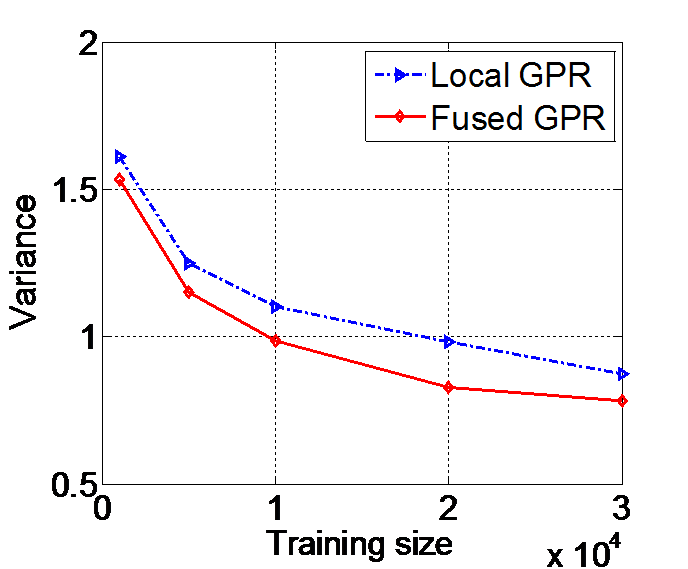}
			\label{Fig:33b}
		\end{minipage}%
	}%
	\subfigure[\emph{Kin40k}]{
		\begin{minipage}[t]{0.5\linewidth}
			\centering
			\includegraphics[width=1.7in,height=1.5in]{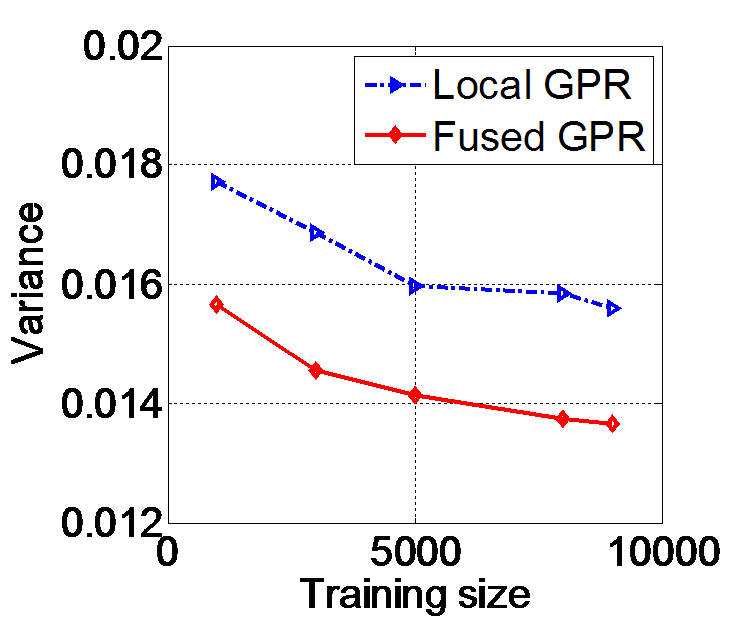}
			\label{Fig:33d}
		\end{minipage}%
	}%
	\centering
	\caption{Performance comparison of the agent-based local GPR and fused GPR on datasets SARCOS and \emph{Kin40k}.}\label{Fig:33}
\end{figure}

We assign $n=100$ agents in the network, and also use the square-exponential kernel $k(\boldsymbol{z},\boldsymbol{z}_*)$. There are five Byzantine agents, i.e., $\alpha=5\%$, and we let $\beta=\alpha$. In this experiment, we {{apply (\ref{eq:99})}}. For the two real-world datasets, we pick agent $76$ and agent $42$, respectively. With the different training data size, we compute the averaged variance for all test points, and Figs. \ref{Fig:33}(a) and \ref{Fig:33}(b) show that the agent-based fused GPR provides a less prediction uncertainty than the agent-based local GPR. This demonstrates that the learning accuracy of the agent-based fused GPR is improved.

\begin{table}[H]
	\caption{Prediction performance comparisons between local GPR and fused GPR on two real-world datasets.}
	\vspace{0.05cm}
	\label{Tab:3}
	\centering
	\begin{tabular}{|c|c|c|c|c|}
		\hline
		Training data size & $10000$ & $20000$ & $30000$ & $40000$ \\
		\hline
		MSE/Sarcos/local & $1.1817$ & $1.0478$ & $0.8669$ & $0.7500$ \\
		\hline
		MSE/Sarcos/cloud & $1.0653$ & $0.8175$ & $0.6954$ & $0.6071$ \\
		\hline
		MSE/Sarcos/fused & $1.0020$ & $0.7564$ & $0.6644$ & $0.5613$ \\
		\hline
		Training data size & $1000$ & $3000$ & $4000$ & $5000$ \\
		\hline
		MSE/\emph{Kin40k}/local & $1.7366$ & $1.6716$ & $1.1605$ & $0.8894$ \\
		\hline
		MSE/\emph{Kin40k}/cloud & $0.8500$ & $0.7388$ & $0.6874$ & $0.6486$ \\
		\hline
		MSE/\emph{Kin40k}/fused & $0.8043$ & $0.7387$ & $0.6701$ & $0.6324$ \\
		\hline
	\end{tabular}
\end{table}

Meanwhile, with training data size increasing, we provide the MSE comparisons between the agent-based fused GPR and the agent-based local GPR in Table \ref{Tab:3}. It can be seen that the prediction error of the agent-based fused GPR is much smaller than that of the agent-based local GPR, {{which implies that the agent-based fused GPR in our algorithm 1 allows the agents to utilize the predictions from the cloud-based aggregated GPR to potentially enhance prediction accuracy by leveraging communication}}.

\section{Conclusion}
We propose a Byzantine-resilient federated GPR algorithm, which is able to tolerate less than one quarter Byzantine agents in the network and improve the learning performance of the agents. We characterize the upper bounds on the prediction errors and the lower and upper bounds of the predictive variances from the cloud-based aggregated GPR. We theoretically quantify the learning performance improvements of the agent-based fused GPR over the agent-based local GPR. Comprehensive experiments are conducted to demonstrate the robustness to Byzantine attacks and learning performance improvements of the Byzantine-resilient federated GPR.

\begin{ack}                               
This work was partially supported by NSF awards ECCS 1846706 and ECCS 2140175.  
\end{ack}

\bibliographystyle{agsm-nq}
\bibliography{autosam}
\begin{autoBiography}
    {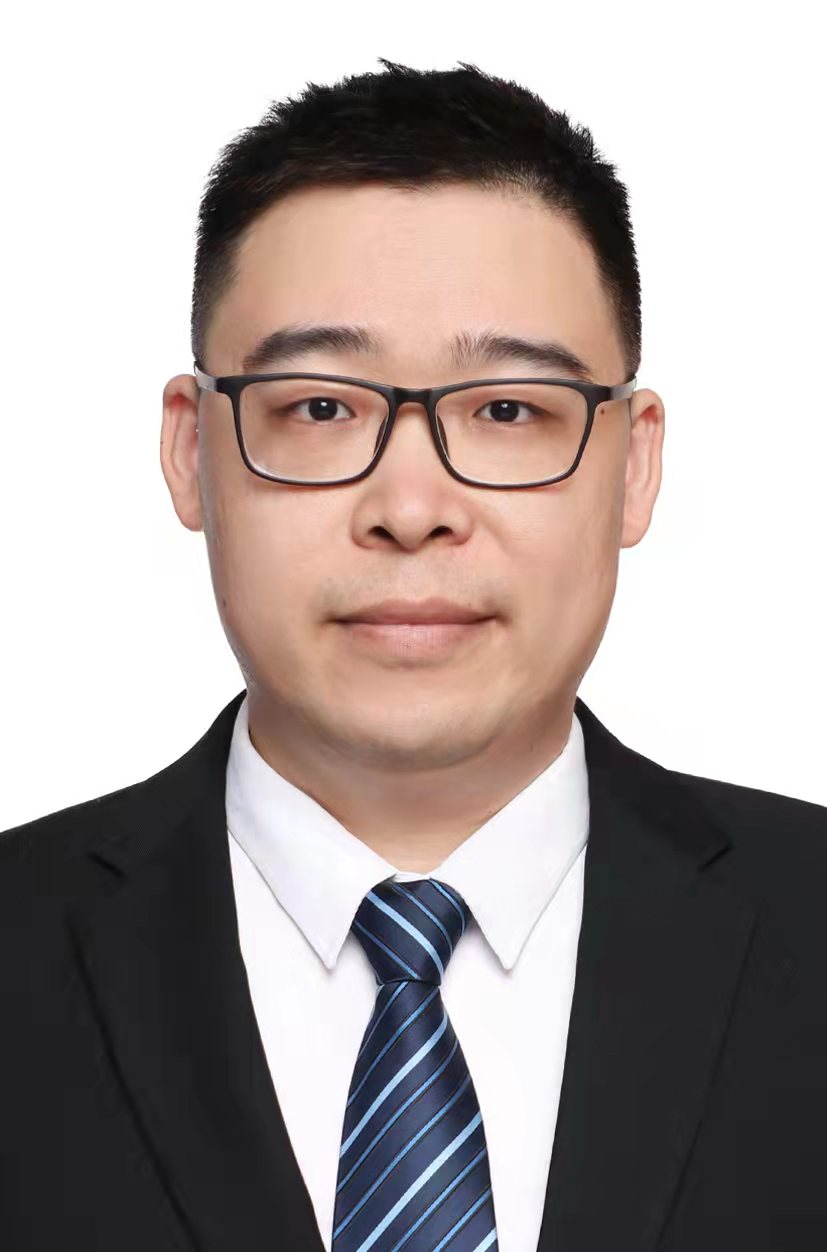}{Xu Zhang}{is a Ph.D. candidate in the School of Electrical Engineering and Computer Science at the Pennsylvania State University. He received B.S. degree in Automation and M.S. degree in Control Science and Engineering from Harbin University and Science and Technology in 2008 and Harbin Institute of Technology in 2010, respectively. From February 2017 to February 2018, he worked as a visiting scholar in the School of Mechanical and Aerospace Engineering at North Carolina State University. His research interests mainly focus on security of cyber-physical systems and machine learning.}
\end{autoBiography}

\begin{autoBiography}
    {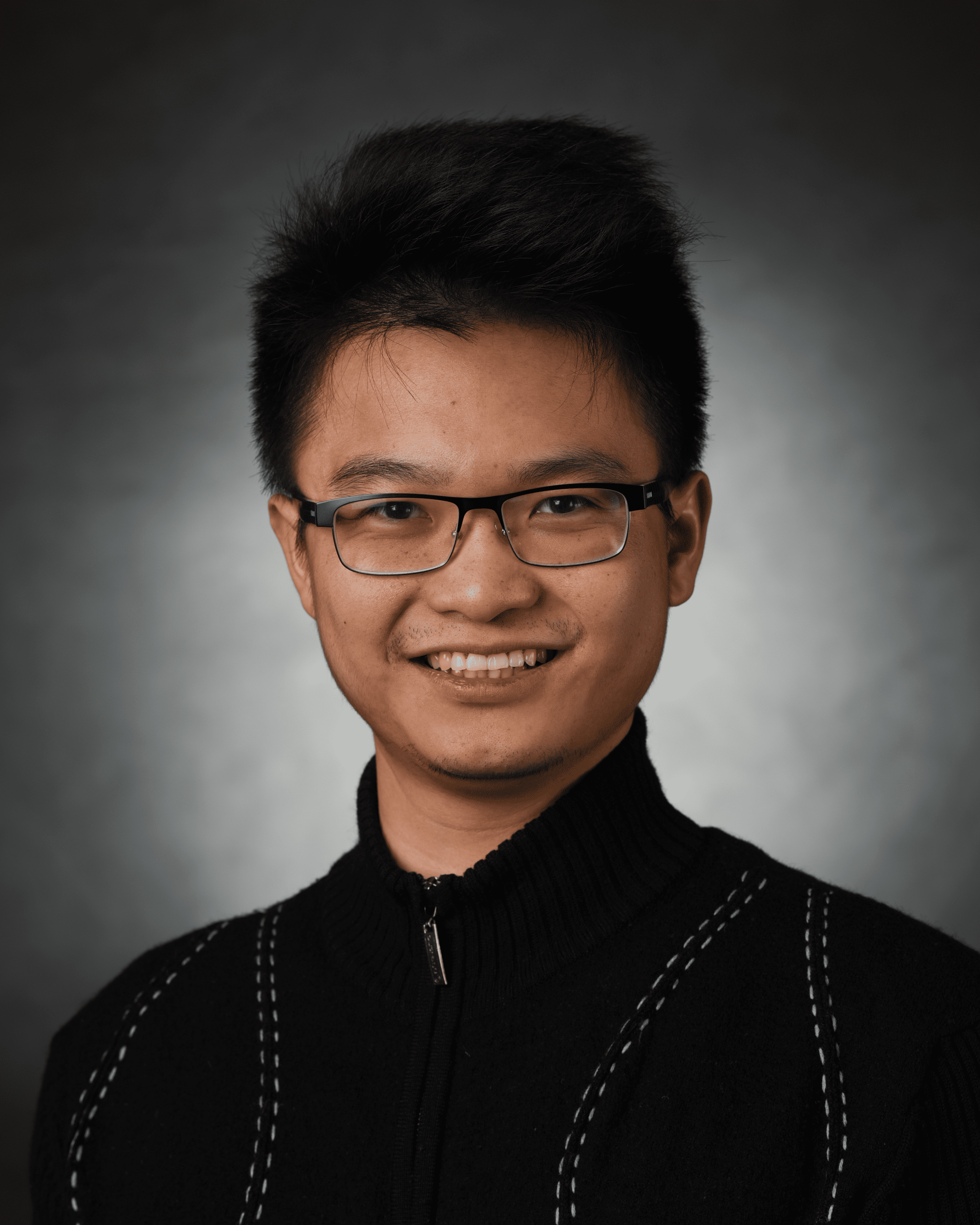}{Zhenyuan Yuan}{is a Advanced Analytics \& Machine Learning Researcher in the Virginia Tech
Transportation Institute at Virginia Polytechnic
Institute and State University (Virginia Tech).
Prior to that, he was a postdoctoral associate in
the Bradley Department of Electrical and Computer Engineering at Virginia Tech. He received
Ph.D. in Electrical Engineering and dual B.S. in
Electrical Engineering and in Mathematics from
the Pennsylvania State University in 2024 and
2018, respectively. His research interests lie in
trustworthy machine learning and motion planning with applications in
multi-robot systems. He is a recipient of the Rudolf Kalman Best Paper
Award of the ASME Journal of Dynamic Systems Measurement and
Control in 2019 and the Penn State Alumni Association Scholarship for
Penn State Alumni in the Graduate School in 2021.}
\end{autoBiography}

\begin{autoBiography}
{Minghui-ZHU.PNG}{Minghui Zhu}{is a Professor in the School of
Electrical Engineering and Computer Science at
the Pennsylvania State University. Prior to joining
Penn State in 2013, he was a postdoctoral
associate in the Laboratory for Information and
Decision Systems at the Massachusetts Institute
of Technology. He received Ph.D. in Engineering
Science (Mechanical Engineering) from the
University of California, San Diego in 2011. He
was a joint appointee as Senior Engineer in
the Optimization and Controls Department at
the Pacific Northwest National Laboratory. His research interests lie
in distributed control and decision-making of multi-agent networks with
applications in robotic networks, security and the smart grid. He is
an author of the book “Distributed optimization-based control of multiagent
networks in complex environments” (Springer, 2015). He is a
recipient of the Dorothy Quiggle Career Development Professorship in
Engineering at Penn State in 2013, the National Science Foundation
CAREER award in 2019, the Superior Paper Award of the American
Society of Agricultural and Biological Engineers in 2024 and the Penn
State Engineering Alumni Society Outstanding Research Award in 2025.
He is an associate editor of Automatica, the IEEE Transactions on
Automatic Control, the IEEE Open Journal of Control Systems and the
IET Cyber-systems and Robotics.}
\end{autoBiography}

\end{document}